\documentclass[acmsmall,screen]{acmart}

\usepackage{epsfig}
\usepackage{graphicx}
\usepackage{amsmath}
\usepackage{epstopdf}
\usepackage{color}
\usepackage{array}
\usepackage{comment}
\usepackage{subcaption}
\makeatletter  
\newif\if@restonecol  
\makeatother

\usepackage{multirow}

\AtBeginDocument{%
  \providecommand\BibTeX{{%
    \normalfont B\kern-0.5em{\scshape i\kern-0.25em b}\kern-0.8em\TeX}}}

\setcopyright{acmcopyright}
\copyrightyear{2018}
\acmYear{2021}
\acmDOI{10.1145/1122445.1122456}

\acmJournal{JACM}
\acmVolume{37}
\acmNumber{4}
\acmArticle{111}
\acmMonth{8}

\acmSubmissionID{2021-0237-R2}

\graphicspath{{./figure/}}
\newcommand{\etal}{\textit{et al.}}
\newcommand{\eg}{\textit{e.g.}}
\newcommand{\ie}{\textit{i.e.}}
\newcommand{\etc}{\textit{etc.}}
\newcommand{\viz}{\textit{viz.}}

\newcommand{\rev}[1]{\textcolor[rgb]{0.0,0.0,0.0}{#1}}
\newcommand{\revrm}[1]{\textcolor[rgb]{0.1,0.1,0.8}{}}

\newcommand{\revii}[1]{\textcolor[rgb]{0.0,0.0,0.0}{#1}}

\newcommand{\mygray}[1]{\textcolor[rgb]{0.2,0.2,0.25}{\scriptsize{(#1)}}}

\begin{document}

\title{Semantic Guided Single Image Reflection Removal}

\author{Yunfei Liu}
\orcid{0000-0001-6898-0058}
\affiliation{%
  \institution{State Key Laboratory of Vitrtual Reality Technology and Systems, School of Computer Science and Engineering, Beihang University}
  \country{China}
  \postcode{100191}
}
\email{lyunfei@buaa.edu.cn}

\author{Yu Li}
\orcid{0000-0003-1865-8276}
\affiliation{%
  \institution{International Digital Economy Academy}
  \city{Shenzhen}
  \country{China}}
\email{yul@Illinois.edu}

\author{Shaodi You}
\orcid{0000-0001-8973-645X}
\affiliation{%
  \institution{University of Amsterdam}
  \city{Amsterdam}
  \country{The Netherlands}
}
\email{s.you@uva.nl}

\author{Feng Lu}
\orcid{0000-0001-9064-7964}
\authornote{Feng Lu is the corresponding author. }
\affiliation{%
	\institution{State Key Laboratory of Vitrtual Reality Technology and Systems, School of Computer Science and Engineering, Beihang University}
	\country{China}
	\postcode{100191}
}
\affiliation{%
\institution{Peng Cheng Laboratory}
\city{Shenzhen}
\country{China}
}
\email{lufeng@buaa.edu.cn}

\renewcommand{\shortauthors}{Liu, et al.}

\setcopyright{acmcopyright}
\acmJournal{TOMM}
\acmYear{2022} \acmVolume{1} \acmNumber{1} \acmArticle{1} \acmMonth{1} \acmPrice{15.00}\acmDOI{10.1145/3510821}

\begin{abstract}
  Reflection is common when we see through a glass window, which not only is a visual disturbance but also influences the performance of computer vision algorithms. Removing the reflection from a single image, however, is highly ill-posed since the color at each pixel needs to be separated into two values belonging to the clear background and the reflection respectively. To solve this, existing methods use additional priors such as reflection layer smoothness, double reflection effect, and color consistency to distinguish the two layers. However, these low-level priors may not be consistently valid in real cases. In this paper, inspired by the fact that human beings can separate the two layers easily by recognizing the objects and understanding the scene, we propose to use the object semantic cue, which is high-level information, as the guidance to help reflection removal. Based on the data analysis, we develop a multi-task end-to-end deep learning method with a semantic guidance component, to solve reflection removal and semantic segmentation jointly. Extensive experiments on different datasets show significant performance gain when using high-level object-oriented information. We also demonstrate the application of our method to other computer vision tasks.
\end{abstract}



\maketitle

\section{Introduction} \label{sec:Introduction}

When taking a photo of objects behind the glass window, unwanted reflection frequently appears, which not only is visually disturbing but may also affect the performance of other computer vision algorithms (\eg, classification~\cite{ImageProcess:liu2020reflection}, object detection~\cite{ImageProcess:yolov3}, scene parsing~\cite{ImageProcess:zhou2017scene}, \etc).  
To solve this problem, reflection removal has been explored by a number of existing works \cite{ ReflectionRemoval:Fan2017A, ReflectionRemoval:Li2013Exploiting, ReflectionRemoval:Nandoriya2017Video,ReflectionRemoval:Wei_2019_CVPR_ERRN,ReflectionRemoval:Wen_2019_CVPR_Linear,ReflectionRemoval:Wieschollek2017Separating}.
Single image reflection removal is a challenging problem, \ie, it only takes one single image as the input and aims to separate it into two outputs, the clear background and the reflection.
More formally, given an input image with reflection, denoted as $\mathbf{I}$, we need to separate it into background $\mathbf{B}$ and reflection $\mathbf{R}$~\cite{ReflectionRemoval:Li2014Single,ReflectionRemoval:Zhang2018PerceptualLosses}:
\begin{equation} \label{equ:Intro}
	\mathbf{I} = \mathbf{B} + \mathbf{R}.
\end{equation}

\begin{figure}[htbp]
	\begin{center}
		\includegraphics[width=1.0\linewidth]{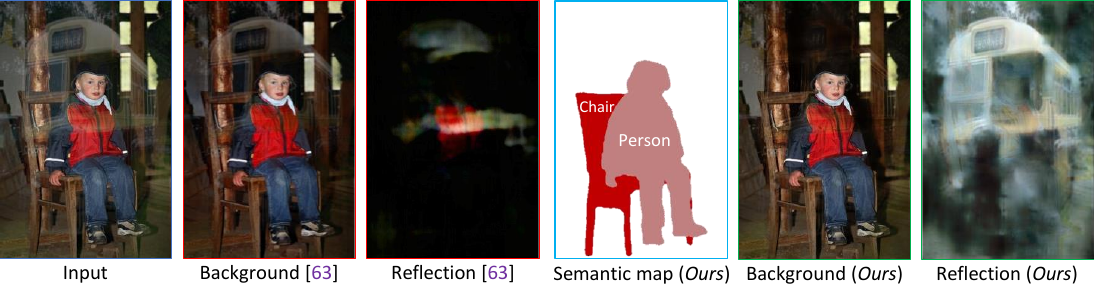}
	\end{center}
	\caption{
		From left to right: The input, background estimation, and reflection estimation from one of the state-of-the-art methods~\cite{ReflectionRemoval:Zhang2018PerceptualLosses}. Our semantic segmentation, background estimation, and reflection estimation. Our approach performs a joint task of reflection removal and semantic segmentation. Understanding the semantics of the background, our methods can provide much clearer separation results.
	}
	\label{fig:semantic_separation}
\end{figure}

Apparently,  Eqn.~\eqref{equ:Intro} is ill-posed due to there are two unknowns (\textbf{B} and \textbf{R}) to be divided from only one known (\textbf{I}). 
Without any constraints, there are numerously possible solutions. Earlier methods propose different priors as constraints (\eg, relative smoothness~\cite{ReflectionRemoval:Li2014Single}, chromaticity consistency~\cite{ReflectionRemoval:Zhao2015SpecularRR}, ghost effect~\cite{ReflectionRemoval:Shih2015Reflection}, \etc) to get a meaningful solution.  
However, such low-level priors are not general enough in real cases. 
Fan~\etal~\cite{ReflectionRemoval:Fan2017A} are the first to train a deep neural network for this task with losses on colors and edges.
Following~\cite{ReflectionRemoval:Fan2017A}, other methods have been proposed in the direction of new network design~\cite{ReflectionRemoval:Wan2017Benchmarking,ReflectionRemoval:Wan2018CRRN,ReflectionRemoval:Wei_2019_CVPR_ERRN,ReflectionRemoval:Wen_2019_CVPR_Linear,ReflectionRemoval:eccv18refrmv,ReflectionRemoval:Zhang2018PerceptualLosses}, new loss functions like perceptual loss~\cite{ReflectionRemoval:Zhang2018PerceptualLosses} and GAN loss~\cite{ReflectionRemoval:Wen_2019_CVPR_Linear,ReflectionRemoval:Ma2019Jointly}, or attention modules \cite{ReflectionRemoval:Wei_2019_CVPR_ERRN}.
All of them use a network to learn a mapping from a blended reflection image to a clear background. 
While adversarial learning using GAN loss and contextual attention may provide information beyond low-level features, the object-level semantic information has never been explored.
We have observed that using only the low-level information (even equips with adversarial learning) is insufficient for the reflection separation due to the high ambiguity in low-level appearance.

Fig.~\ref{fig:semantic_separation} shows a case, though \cite{ReflectionRemoval:Zhang2018PerceptualLosses} introduced high-level constraints, they still cannot handle this case well. 
In this paper, we use a simple intuition from human cognition, which we humans can easily separate the mixed visual appearance into reflection and background layers by object level priors. 
One can first recognize the internal objects and quickly assign them to the background layer. Take Fig.~\ref{fig:semantic_separation} as an example, we understand that the background is a person sitting on a chair so that all human parts and the chair need to be in the background layer. This enables us to recover the background correctly and meanwhile recover the reflection layer clearly.

However, implementing such an idea is not trivial, because it is not guaranteed to get the robust semantics from the input image with reflection, while a cleaner image is of many benefits to the estimation of semantic segmentation. To enable this, we assume the intensity of the background image is stronger than the reflection, which is the same as all existing reflection removal methods. Therefore, the background's semantics is more significant than that in the reflection layer in this situation.
We then propose a novel multi-task learning method for single image reflection removal, named Semantic Guided Reflection Removal Network (SGR$^2$N). This means the semantic estimation and reflection removal are learned and optimized simultaneously.
Furthermore, we design the semantic guidance block, which explicitly utilizes the semantic to guide the reflection removal. 

To evaluate the effectiveness of SGR$^2$N, we conducted systematical experiments on three datasets.
Experiments report consistent and significant performance improvement on all three datasets. Ablation study and analytical experiments also show the effectiveness of our method.

\textbf{Contributions.} We summary the contributions as follows:

\begin{itemize}
	\item To the best of our knowledge, \rev{it is the first method to use pixel-level semantic information as guidance for reflection removal, which is an explicit way to represent and use semantic information. And our method jointly solves the semantic segmentation and reflection removal from a single image.}
	
	\item We propose a novel multi-task and end-to-end network with a novel semantic guidance block to achieve single simultaneous image reflection removal (main-task) and semantic analysis (sub-task). 
	
	\item We demonstrate the effectiveness of the method through comprehensive experiments on both three datasets.   
\end{itemize}

\section{Related Work}

\noindent\textbf{Multiple-input methods.} 
Many works solve Eqn.~\eqref{equ:Intro} with multiple inputs. Methods~\cite{ReflectionRemoval:Sinha2012Image, ReflectionRemoval:Guo2014Robust,ReflectionRemoval:Li2013Exploiting,ReflectionRemoval:Simon2015Reflection,ReflectionRemoval:Nandoriya2017Video} assume the reflection and background layer are at different depth planes which can be separated by multi-view depth estimation.
To align multiple inputs, optical flow is adopted to achieve reflection removal \revii{\cite{ReflectionRemoval:Xue2015A, ReflectionRemoval:Yang2016Robust,ReflectionRemoval:liu2020learning}}. 
Another group of work uses a pair of with/without flash points to make reflection removal like~\cite{ReflectionRemoval:Agrawal2005Removing}. 
Schechner \etal \cite{ReflectionRemoval:Schechner1998Separation} using a group of images with different focal lengths, remove reflections by solving the depth 
of different layers. 
Punnappurath~\etal \cite{ReflectionRemoval:punnappurath2019reflection} use dual pixel sensors to capture two sub-aperture views of the scene, and then find defocus disparity clues for reflection removal.
\revii{Many works \cite{ReflectionRemoval:Kong2014A, ReflectionRemoval:Wieschollek2017Separating,ReflectionRemoval:Lei2020Polarized,ReflectionRemoval:li2020reflection,ReflectionRemoval:wen2021polarization}} explore the polarization and take multiple images to solve the optimal separation through angle filter.
\rev{Lei \etal \cite{ReflectionRemoval:lei2021robust} propose to use a pair of flash and no-flash images as input. Then they exploit flash-only cues for reflection removal.
Niklaus \etal \cite{ReflectionRemoval:niklaus2021learned} propose a learning-based dereflection algorithm that uses stereo images as input. They find the cues for reflection removal from two views.}

\noindent\textbf{Non-CNN single-image methods.}
Eqn.~\eqref{equ:Intro} is not directly solvable for a single image.	To tackle this 
Li \etal \cite{ReflectionRemoval:Li2014Single} assume that the reflection layer is more blurry than the background layer and model these as two different gradient distributions in the two layers for the separation. 
Shih \etal \cite{ReflectionRemoval:Shih2015Reflection} explore the ghost effect in the reflection layer and design a GMM model to make reflection removal.
Arvanitopoulos \etal \cite{ReflectionRemoval:Arvanitopoulos2017Single} make reflection suppression through the relative gradient prior between two different layers. 
Sandhan \etal \cite{ReflectionRemoval:Sandhan2017Anti} use the symmetry in the human face to remove the reflections on glasses. 
Yun \etal \cite{ReflectionRemoval:Yun2018Reflection} propose an algorithm to remove virtual points in large-scale 3D points clouds using the conditions of the reflection symmetry and the geometric similarity.
Yang \etal~\cite{ReflectionRemoval:Yang_2019_CVPR_Suppression} propose a relatively fast reflection suppression algorithm via convex optimization. 

\noindent\textbf{CNN-Based single image methods.} 
Fan \etal \cite{ReflectionRemoval:Fan2017A} propose the Cascade Edge and Image Learning Network (CEIL Net) for reflection removal, in which the background's edge is predicted at first and then is used to guide the reflection separation.
Wan \etal utilize existing prior information, design a benchmark \cite{ReflectionRemoval:Wan2017Benchmarking} for reflection removal, and then train an end-to-end model called CRRN \cite{ReflectionRemoval:Wan2018CRRN} to separate layers. \rev{Wan \etal \cite{ReflectionRemoval:wan2019corrn} adopt multi-scale gradient information for reflection removal.}
\rev{Inspired from image-to-image translation, Liu \etal \cite{ReflectionRemoval:liu2020separate} proposed an unsupervised method for reflection removal. Kim \etal \cite{ReflectionRemoval:kim2020unsupervised} extend the deep image prior to unsupervised reflection removal. }
Zhang \etal \cite{ReflectionRemoval:Zhang2018PerceptualLosses} propose perceptual loss, which is extracted from the first layers of VGG, later they combine feature loss, adversarial loss, and exclusive loss together.
The main difference between perceptual loss 
and ours is that we explicitly utilize high-level semantic information to guide 
reflection removal during training. 

To suit the method for more challenging scenarios, Wen \etal \cite{ReflectionRemoval:Wen_2019_CVPR_Linear} propose to train one CNN to synthesize the reflection contaminated image beyond linearity.
Wei \etal \cite{ReflectionRemoval:Wei_2019_CVPR_ERRN} train ERRNet to remove reflections from little misaligned image pairs.
More recently, Ma~\etal~\cite{ReflectionRemoval:Ma2019Jointly} introduce the entanglement and disentanglement mechanisms to reflection removal.
\rev{Hong~\etal~\cite{ReflectionRemoval:Hong_2021_CVPR} study the reflection removal on panoramic image, which includes a partial view of reflection scenes. Zheng~\etal~\cite{ReflectionRemoval:zheng2021single} explore the absorption effect and design a two-stage framework for reflection removal.}
There are also many methods that try to tackle reflection removal in a multi-stage manner~\cite{ReflectionRemoval:eccv18refrmv,ReflectionRemoval:Li2020Cascaded}.
Differ from the aforementioned methods, 
\rev{we explicitly use the pixel-level semantic information for reflection removal task. Specifically, we adopt intermediate features from the semantic segmentation task, which contain meaningful object attentions, as guidance signals for reflection removal.}

\noindent\textbf{Semantic information in other low-level vision tasks.}
Semantic information has been used in many low-level vision tasks. For instance, 
Baslamisli \etal~\cite{ImageProcess:baslamisli2018joint} propose to use semantic information to guide albedo computation and design cascade CNNs for joint learning of intrinsic images and semantic segmentation.
Liu \etal~\cite{ImageProcess:liu2020connecting} propose a deep neural network solution that cascades two modules for image denoising and high-level tasks, respectively.
Wen \etal~\cite{ImageProcess:ren2018deep} propose to incorporate global semantic priors as input to regularize the image dehazing. 
Wang \etal~\cite{ImageProcess:wang2020cascaded} propose a three-stage model for rainy image restoration. In which a semantic consistency loss is adopted for details recovery. 
These methods usually adopt multi-stage designs to estimate the semantics and low-level vision task successively. Unlike these problems, we proposed a novel unified framework with a semantic guidance module that jointly learns semantic segmentation and reflection removal.


\section{Semantic Guided Reflection Removal}

\subsection{Study on Semantic Information with Reflection Interference} \label{sec:valid_analysis}

\begin{figure}[htbp]
	\begin{center}
		\includegraphics[width=0.7\linewidth]{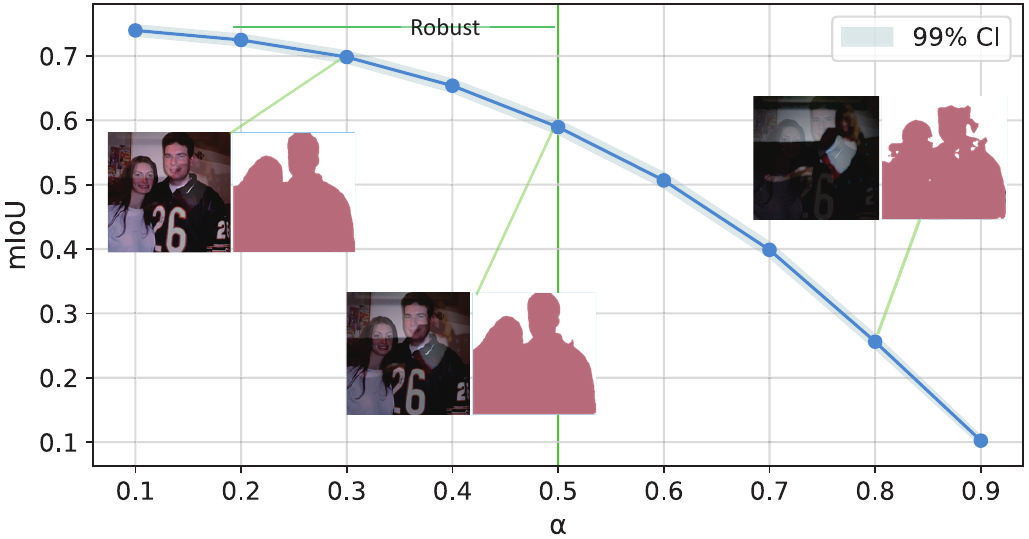}
	\end{center}
	\caption{Relationship between mIoU
		(metric for semantic segmentation) and the weight ($\alpha$) of reflection layer with one visual example. 
		Here CI is the confidence interval. 
		We find that semantic estimation is robust in the observations with a relatively low weight of the reflection layer. }
	\label{fig:sepatation_unstable}
\end{figure}

Our motivation is that if the background layer semantic information is provided or can b estimated, we can use it to help separate the background layer from the input image.  
In other low-level vision tasks to recover the clean image from the interference, the interference layer to be removed is usually largely different from the image layer and contains no semantic object information, \eg, noise, hazy volume, rain streaks. Unlike these tasks, in reflection removal, the reflection layer to be removed is also a reflected scene, which causes difficulty in estimating the background semantics. Therefore, it is important to see how the reflection affects the estimation of background semantics.

We try an empirical study to test the robustness of semantic segmentation estimation against different intensity levels of reflection.
We randomly sample images from the Pascal VOC dataset \cite{Semantic:Everingham2010pascalVOC}, where the ground truth of semantic label is provided in 20 categories. 
Based on it, we synthesize the image with reflections by linearly blending two images using $\mathbf{I} = (1-\alpha)\mathbf{B} + \alpha \mathbf{R}$, 
where larger $\alpha$ can simulate the reflection layer with stronger intensity.
In total, we sample and generate $5000 \times 9$ sets of images with $\alpha=0.1, 0.2, \dots, 0.9$.
Fig.~\ref{fig:sepatation_unstable} illustrates the relationship between semantic segmentation quality and the weight of reflection layer. 

Statistical results have shown that semantic estimation is relatively robust when
the weight of reflection layer is below 50\%. 
When the weight is bigger than 50\%, such reflections would be taken as extreme cases, which might contain 
1) reflection layer is too strong or 
2) background layer contains no salient object (and leads texture-less transmission layer). 
In this work, following all previous methods ~\cite{ReflectionRemoval:Fan2017A,ReflectionRemoval:Zhang2018PerceptualLosses}, we make a similar assumption and address the cases with moderate reflection strength and the background.
The extreme cases with intense reflections are considered as limitations of our method and leave such cases as our future directions.

\subsection{Multi-task Learning for Simultaneous Reflection Removal and Semantic Estimation}

\begin{figure*}[htbp]
	\begin{center}
		\includegraphics[width=\textwidth]{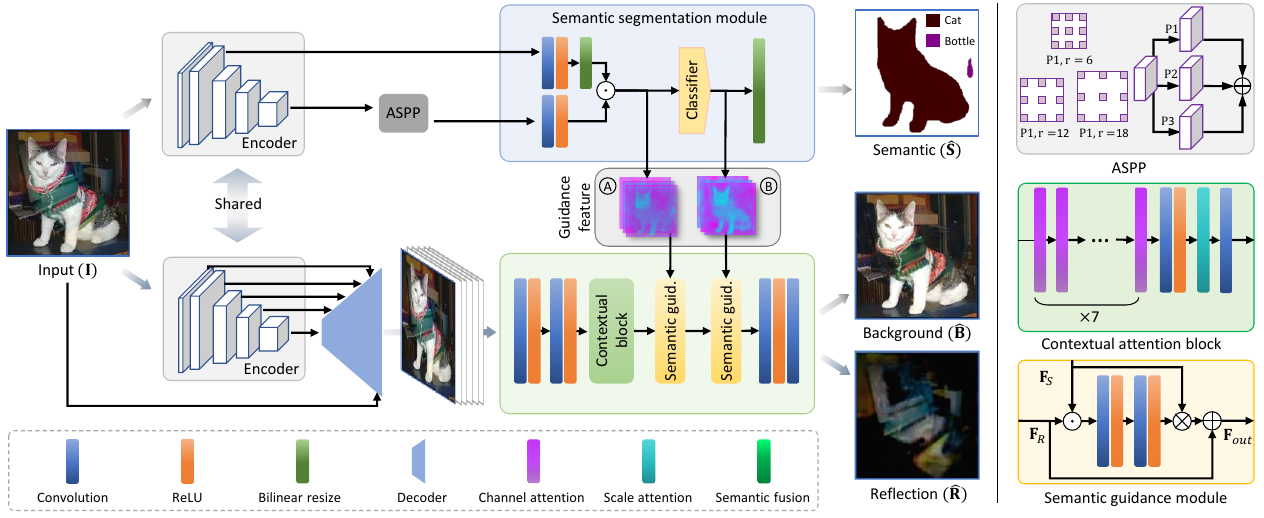}
	\end{center}
	\caption{An overview of our proposed SGR$^2$N. \textbf{Left:} For the input image \textbf{I}, we first extract features via the Shared Encoder, and estimate the semantic map $\mathbf{S}_B$ of the background through \textit{Semantic segmentation} module. Next, the semantic information is used to guide the \textit{Reflection separation} through a semantic guidance module. Finally, background (\textbf{B}) and reflection (\textbf{R}) are estimated from the separation module. \textbf{Right:} \rev{From top to down: the architecture of ASPP module, contextual attention block, and semantic guidance module.} }
	\label{fig:main_figure}
\end{figure*}

Given an input image with reflection interference, we perform two tasks: (1) Extracting background
semantic map $\mathbf{S}_B$ from input $\mathbf{I}$, and 
(2) Recovering background layer $\mathbf{B}$
(also reflection layer $\mathbf{R}$) from input $\mathbf{I}$ along with the semantic information ($\mathbf{S_B}$) obtained in the first task.
To this end, we propose the semantic guide reflection removal network (SGR$^2$N), in which a novel semantic guidance module is proposed to link semantic segmentation and reflection removal together. Our method can work adaptively using semantic information as guidance for reflection removal. 

\subsubsection{Architecture design.}

Our SGR$^2$N layout is illustrated in Fig.~\ref{fig:main_figure}, containing a \rev{\textit{Shared Encoder}}, a \textit{Semantic segmentation} module, and a \textit{Separation} module with \textit{Semantic guidance}. 
Given the input image, \textit{Shared Encoder} aims to extract rich features from different stages of the backbone \rev{for semantic segmentation task and reflection removal task}. \rev{Following \cite{Semantic:Chen2018DeepLabV3plus}, we use ResNet-101~\cite{Semantic:He2015Deep} as the backbone}, in which intermediate features from different stages \revrm{(res 1-5)} are taken as outputs.
\textit{Semantic segmentation} module estimates the semantic map of the input. We follow the DeepLabv3+~\cite{Semantic:Chen2018DeepLabV3plus} and move its ASPP module to the decoder. We use the semantic classifier to predict the semantic map of the input image. 
\rev{The semantic classifier is constructed by a $1\times 1$ convolutional layer and a softmax activation layer. It maps the feature into the semantic predictions (with the shape of $[B, N, H, W]$, where $N$ is the number of semantic labels).}
The intermediate features before/after the classifier are used as guidance for reflection separation.  

\textit{Separation module} utilizes the fused features as input and tries to recover the corresponding background and reflection with semantic guidance. In this module, we use contextual attention to get richer information firstly. This module is proved effective in low-level vision tasks in \cite{ReflectionRemoval:Wei_2019_CVPR_ERRN}. We bring a similar architecture from \cite{ReflectionRemoval:Wei_2019_CVPR_ERRN} but with fewer channel attention layers.	

\textit{Semantic guidance for reflection removal.} 
We carefully design a semantic guidance module to use semantic information as guidance for reflection removal.
As shown in Fig.~\ref{fig:main_figure}, the detailed design is as follows.
(1) Selection of guidance features. We choose the mid-level feature is because it contains the most informative features from the semantic segmentation module. \revrm{Specifically, there are two intermediate features (\ie, denoted as \textcircled{\small{A}} and \textcircled{\small{B}} in Fig.~\ref{fig:main_figure}) that are used for guiding the separation network.} \rev{As illustrated in Fig.~\ref{fig:main_figure}, we adopt the intermediate features (\textcircled{\small{A}} and \textcircled{\small{B}}) of semantic segmentation module. These intermediate features ensure the guidance contains the soft object's attentions for reflection removal. In detail,} 
	a) \textcircled{\small{A}} contains all information about high-level information (generated by ASPP module) and low-level feature (from the output the second residual block of the encoder). 
	b) \textcircled{\small{B}} contains all the features of the segmentation result.
(2) Fusing strategy of guidance. The underlying design principle is that we use semantic information to make the network focus on meaningful objects for reflection separation. One of the best ways is to use an attention mechanism to introduce semantic information across channels~\cite{Semantic:hu2018squeeze}. We use the commonly used attention operation for implementation. Based on the standard attention layer, we further get a richer structure by making it within a residual block, which can also avoid a dramatic increase of the trainable parameters. 

\rev{Here we use two semantic guidance blocks to use mid-level and high-level semantic features to guide reflection removal.}
Specifically, let $\textbf{F}_R$ denote original features produced by the former network layer, $\textbf{F}_S$ denote the semantic feature map produced by the semantic estimation module. $\textbf{F}_S$ mainly provides contextual attention to the image. Mathematically, the semantic guidance module uses semantic feature $\textbf{F}_S$ as guidance for original features $\textbf{F}_R$ as follows:
\begin{equation} \label{equ:semantic_fusion}
		\textbf{F}_{out} = \textbf{V} \otimes \textbf{F}_S \oplus \textbf{F}_R, \quad
		\textbf{V} = \tau(\textbf{w}_{v1}*\tau(\textbf{w}_{v2}*\textbf{K} + \textbf{b}_{v2}) + \textbf{b}_{v1}), \quad
		\textbf{K} =  \textbf{F}_R \odot \textbf{F}_S,
\end{equation}
where $\tau$ is ReLU activation function, $\textbf{w}_{v1}, \textbf{b}_{v1}, \textbf{w}_{v2}, \textbf{b}_{v2}$ are learnable parameters.
For the regions with no semantic information, the semantic guidance module will yield the $\textbf{F}_R$ directly thanks to the residual structure. This design can provide relatively robust guidance for separation.
 
The output layer of the reflection removal module is a convolution layer with 6 channels. The former 3 channels construct the background layer $\textbf{B}$, and the remaining channels construct the reflection layer  $\textbf{R}$.

\subsubsection{Loss function}

As we are jointly performing two tasks, the final multi-task learning (MTL) loss functions are built on reflection removal task and semantic segmentation together.

\textit{Reflection removal loss.} We use three terms of loss functions for regularizing $\mathcal{L}_B$ and $\mathcal{L}_R$, which is similar to previous methods~\cite{ReflectionRemoval:Wei_2019_CVPR_ERRN,ReflectionRemoval:eccv18refrmv,ReflectionRemoval:Zhang2018PerceptualLosses}.

1) Feature loss. Inspired by \cite{ReflectionRemoval:Zhang2018PerceptualLosses,ImageProcess:rad2019srobb}, we define the feature loss based on the features from the \textit{Rich Encoder}. This design will not introduce extra resources. Take image $\textbf{X}$ as the input of encoder, let $\phi_l$ being the feature from the $l$-th stage of \textit{Rich Encoder}, the feature loss is defined as: 
\begin{equation} \label{equ:loss_feat}
	l_{feat} = \sum_l \lambda_l\parallel \phi_l(\textbf{X}) - \phi_l(\hat{\textbf{X}})\parallel_1,
\end{equation}
where $\{\lambda_l\}$ denotes balancing weight. 

2) Pixel loss. Following~\cite{ReflectionRemoval:Fan2017A,ReflectionRemoval:Wei_2019_CVPR_ERRN}, we compute the differences between the prediction and the ground-truth in pixel-level and gradient level as follow:
\begin{equation} \label{equ:loss_pixel}
	l_{pix} = \parallel\textbf{X} - \hat{\textbf{X}}\parallel_1 + \parallel\nabla\textbf{X} - \nabla\hat{\textbf{X}}\parallel_1,
\end{equation}
where $\nabla$ is the gradient operator. \rev{ $\hat{\textbf{X}}$ and $\textbf{X}$ are the network prediction and corresponding ground-truth.}

3) Adversarial loss. We further add adversarial loss to make the produced background and reflection look more realistic. We use a discriminator network $D_\theta$ and minimize the adversarial loss~\cite{I2I:MUNIT}, which is defined as 
\begin{equation} \label{equ:loss_adv}
	l_{adv} = -\log (D_\theta(\textbf{X}, \hat{\textbf{X}})),
\end{equation}
where the loss for discriminator network is $l_{adv}^D = -\log (1- D_\theta(\textbf{X}, \hat{\textbf{X}})) - \log (D_\theta (\textbf{X}, \hat{\textbf{X}}))$.

To summarize, the loss for background ($\mathcal{L}_B$) and reflection ($\mathcal{L}_R$) are
\begin{equation} \label{equ:rr_loss}
	\begin{aligned}
		\mathcal{L}_B &= w_1 l_{feat} (\hat{\textbf{B}}, \textbf{B}) + w_2 l_{pix} (\hat{\textbf{B}}, \textbf{B}) + w_3 l_{adv} (\hat{\textbf{B}}, \textbf{B}), \\
		\mathcal{L}_R &= w_1 l_{feat} (\hat{\textbf{R}}, \textbf{R}) + w_2 l_{pix} (\hat{\textbf{R}}, \textbf{R}) + w_3 l_{adv} (\hat{\textbf{R}}, \textbf{R}), \\
	\end{aligned}
\end{equation}
where we empirically set $w_1=0.1$, $w_2=1$ and $w_3=0.01$ in our experiments. 

\textit{Semantic segmentation loss.} 
For semantic loss $\mathcal{L}_S$, we use cross-entropy as the penalization.
\begin{equation}\label{equ:loss_semantic_new}
	\mathcal{L}_{S} = \sum_{j=1}^{K}\sum_{i=1}^{M}\left(-y_{ji}\log{\hat{y}_{ji}} - \left(1-y_{ji}\right)\log{\left(1-\hat{y}_{ji}\right)}\right),
\end{equation}
where $K$ is the size of training batch, $M$ is the \revrm{summation over classes} \rev{total number of semantic categories. $M=21$ in our synthetic dataset}, $\hat{y}$ is the prediction, the ground truth label is $y$.	

\textit{Total loss.} 
The total loss is the combination of reflection removal loss and segmentation loss.
\begin{equation} \label{equ:loss_mtl}
	\mathcal{L} = \alpha \mathcal{L}_B + \beta \mathcal{L}_R + \mathcal{L}_S,
\end{equation}
where the hyper-parameters are empirically set as $\alpha=0.5$, $\beta=0.25$.

\rev{\subsubsection{Implementation details.} The proposed method is implemented via PyTorch~\cite{ML:PyTorch}, which is available at Github\footnote{\url{https://github.com/DreamtaleCore/SGRRN}}. \rev{To make our model converge faster and save training time,} we employ ResNet-101 with pre-trained parameters on ImageNet \cite{Semantic:imagenet_cvpr09}, \rev{whose effectiveness has been proved in \cite{Semantic:zhou2014learning}.}
Convolution weights are initialized as CZ18~\cite{Semantic:Chen2018DeepLabV3plus}. Input images are resized to $256\times 256$ before feeding into the network.
We train the SGR$^2$N with 500 epochs using the Momentum optimizer \cite{ML_Theory:Sutskever2013Momentum}, which is employed with $\text{momentum} = 0.99$, where the cycle learning rate is initially set as 0.007 and decay in every 30000 iterations until 0.0001. The batch size during training is empirically set to 5.}

\section{Experiments} \label{sec:experiments}

We first synthesis the semantically annotated dataset. 
Then we perform rigorous ablation studies to analyze the effect of different parts of the SGR$^2$N. We evaluate our semantic guided reflection removal method quantitatively and qualitatively on single image reflection removal against previous methods. Next, we make additional experiments on how the intensity of reflection layer affects the final performance of the semantic segmentation and the reflection removal task. Finally, we show applications of our model and make a discussion on failure cases.

\subsection{Semantic Reflection Dataset Generation}  \label{sec:training_data_gen}

\begin{figure}[htbp]
	\begin{center}
		\includegraphics[width=\linewidth]{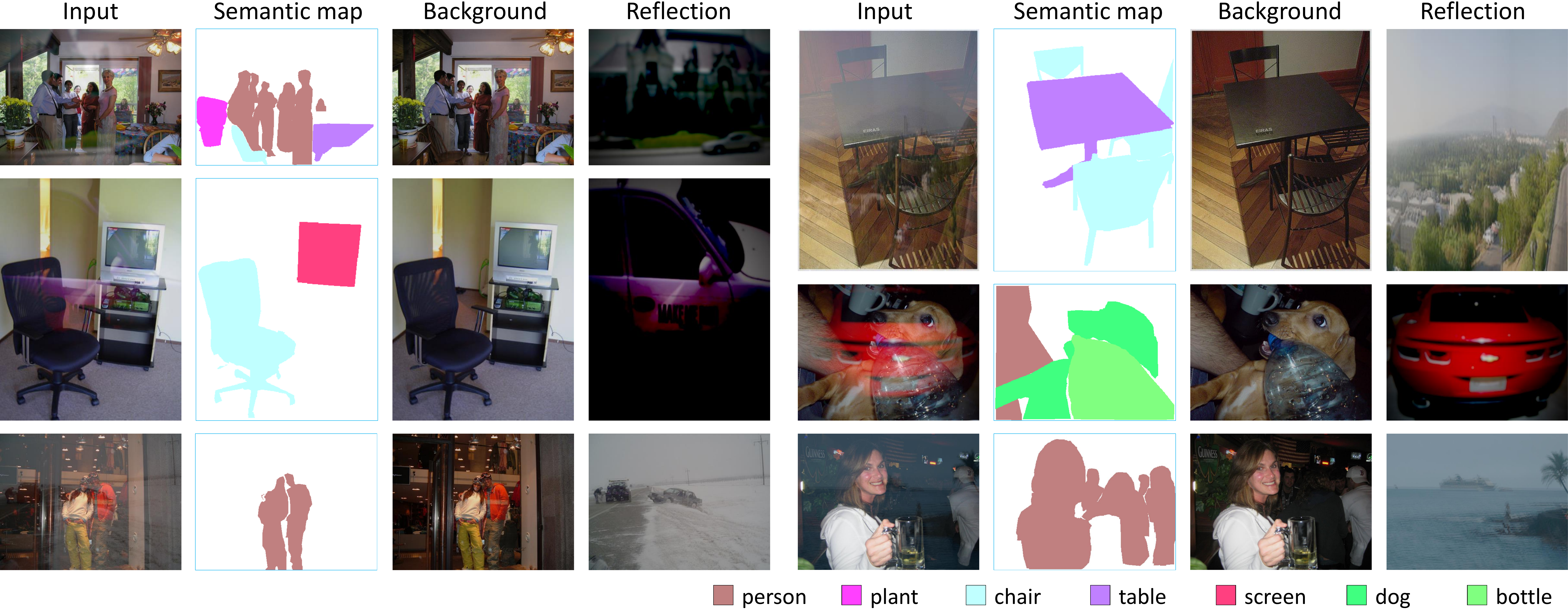}
	\end{center}
	\caption{Illustration of the generated semantic reflection dataset. There are abundant annotations with different semantic meanings for each input image. Here we show three input images with over-smoothness reflection (left-1, left-2, and right-2), the other input images are contaminated by reflection with ghost effect.}
	\label{fig:dataset_visual}
\end{figure}

To train the SGR$^2$N, a training set with semantic annotations is needed. However, existing datasets~\cite{ReflectionRemoval:Fan2017A,ReflectionRemoval:Wan2017Benchmarking,ReflectionRemoval:Zhang2018PerceptualLosses} for reflection removal are all lack semantic annotations. On the other hand, it's labor-consuming to label semantic annotations for the existing reflection removal dataset. Therefore, we generate the reflection removal dataset with semantic labels. 
In detail, there are four types of aligned data in each group of the proposed dataset: (1) image with reflection $\mathbf{I}$, (2) clear background  $\mathbf{B}$, (3) reflection $\mathbf{R}$, and most importantly (4) semantic labels $\mathbf{S}_B$.

To render the realistic reflections, we follow the generation strategy in \cite{ReflectionRemoval:Zhang2018PerceptualLosses}, besides, we add ghost effect~\cite{ReflectionRemoval:Shih2015Reflection} for more abundant and realistic cases.
For semantic labels, we generate the dataset on the basis of Pascal VOC~\cite{Semantic:Everingham2010pascalVOC}, which provides semantic annotations. 
The visual samples of the generated dataset are illustrated in Fig.~\ref{fig:dataset_visual}. 
The generated dataset contains 20 categories of objects, and each image contains 2.93 foreground objects with semantic labels on average. Additionally, the comparison with the existing datasets is listed in Table~\ref{tab:brief_of_all_dataset}.	

\begin{table}[htbp]
	\begin{center}
		\caption{Brief comparison among different datasets.}
		\label{tab:brief_of_all_dataset}
		\setlength{\tabcolsep}{2mm}
		\begin{tabular}{lccccc}
			\toprule[1.3pt]
			Dataset & Source & Volume & Reflection layer & Ghost cases & Semantic label \\
			\hline
			\specialrule{0em}{1pt}{1pt}
			$\mathcal{D}_{BKL}$ & Zhang \etal~\cite{ReflectionRemoval:Zhang2018PerceptualLosses} & 110 &  &  &  \\
			$\mathcal{D}_{SIR^2}$ & Wan \etal~\cite{ReflectionRemoval:Wan2018CRRN} & 454 & $\checkmark$ &  $\checkmark$ &  \\
			$\mathcal{D}_{syn}$ & Ours & 31965 & $\checkmark$ & $\checkmark$ & $\checkmark$ \\
			\bottomrule[1.3pt]
		\end{tabular} 
	\end{center}
\end{table}

\subsection{Experimental Setups}

We train the SGR$^2$N on our synthetic training set, and then we evaluate the performance on the different test sets. Specifically, we use 29503 images from our generated dataset as the training data. Then we evaluate the trained SGR$^2$N on the synthetic test set, 20 real images from Berkeley dataset~\cite{ReflectionRemoval:Zhang2018PerceptualLosses} and 463 real images from SIR$^2$ dataset~\cite{ReflectionRemoval:Wan2018CRRN}.
Image is cropped with the size of 224 $\times$ 224 randomly and is flipped randomly as data augmentation. For the compared methods, we use the same dataset for training, unless specifically stated. For evaluation, we use PSNR, SSIM~\cite{ImageBase:Zhou2004SSIM} to measure the quality of reflection removal. 
\rev{Although these two metrics are most widely used in low-level vision tasks, they have their limitations during the test. It is mainly because SSIM and PSNR are both sensitive to the intensity variance. After all, they leverage luminance and contrast similarity between two images. Therefore, we further use histogram alignment techniques to better calculate SSIM and PSNR in Sec.4.4.}
\rev{Following recent and popular semantic segmentation papers~\cite{Semantic:Chen2018DeepLabV3plus,Semantic:carion2020end}}, we use mean Intersection over Union (mIoU) and \rev{pixel accuracy} of different categories for semantic segmentation. 

\subsection{Ablation Study} \label{sec:ablation_study}

In this section, we evaluate the effectiveness of 1) our multi-task learning architecture, analyze the 2) semantic guidance module and 3) loss functions.

\begin{figure}[htbp]
	\begin{center}
		\includegraphics[width=0.9\linewidth]{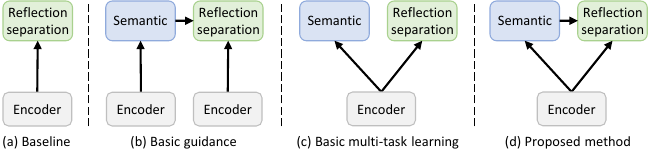}
	\end{center}
	\caption{Different settings of the semantic guided reflection removal model. (a) Train semantic segmentation task solely. (b) The basic multi-task learning model in which two tasks share the same encoder. (c) Our final model in which semantic estimation branch shares guidance to reflection removal branch. 
	}
	\label{fig:MTL_model_choose}
\end{figure}

\textbf{Architecture.}
To evaluate the effectiveness of our semantic guidance architecture, we remove the multi-task scheme or semantic guidance block and re-train the network under the same condition. We first reconstruct our model with training reflection separation solely and take this version as the baseline, as shown in Fig~\ref{fig:MTL_model_choose} (a). In this case, the architecture of baseline is similar to previous state-of-the-art methods~\cite{ReflectionRemoval:Zhang2018PerceptualLosses,ReflectionRemoval:Wei_2019_CVPR_ERRN}.

\begin{figure}[htbp]
	\begin{center}
		\includegraphics[width=\textwidth]{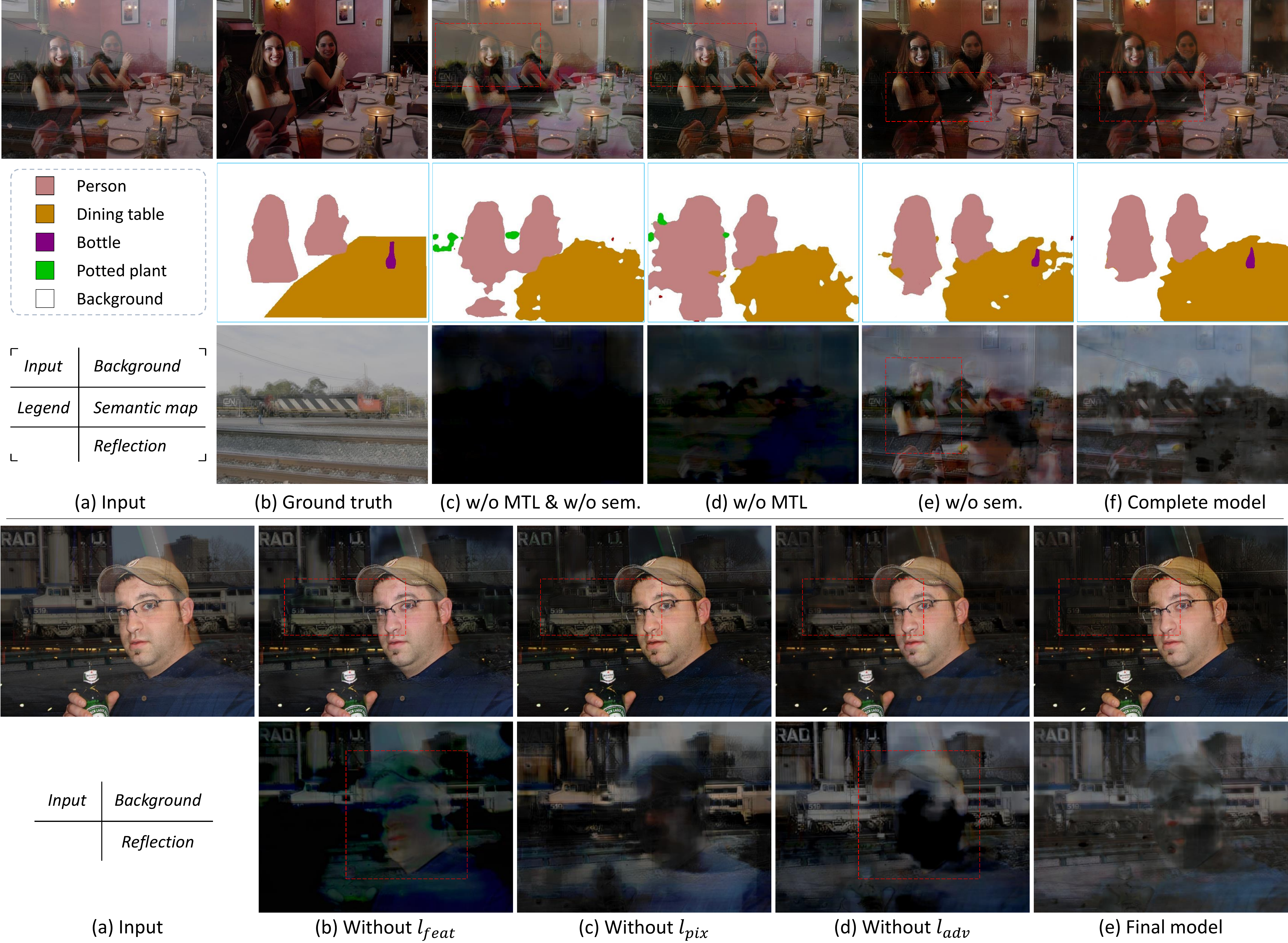}		
	\end{center}
	\caption{
		\rev{Visual results of ablation studies. Top: Visual comparisons of reflection removal results with different network structures. Bottom: Visual results about the loss functions of the SGR$^2$N. }
	}
	\label{fig:ablation}
\end{figure}

We find that reflection cannot be removed in Fig.~\ref{fig:ablation}-top (c), and even unpleasant color-shift appears at the area of reflections. Meanwhile, as shown in the \revrm{second}\rev{third} row of Fig.~\ref{fig:ablation}-top (c), the reflection layer separated from the input contains nothing meaningful. The reason is that the baseline is a reflection removal network with no semantic awareness. 

\textit{Semantic guidance block.} 
To evaluate the effectiveness of the semantic guidance block, we conduct additional experiments by adding semantic guidance into the baseline, as illustrated in Fig.~\ref{fig:MTL_model_choose} (b).
Specifically, we train the semantic segmentation network at first, then freeze the semantic segmentation branch and use its decoder to guide reflection separation through semantic guidance block.
Table~\ref{tab:ablation_semantic_guid} shows the ablation study for architecture. This operation can improve the performance of reflection separation (background/reflection) by $3.03/0.58$ PSNR and $0.007/0.035$ SSIM. The visual results at the top of Fig.~\ref{fig:ablation} (d) show that parts of reflection can be separated from the input, but there are still many residuals that remain on the girl's faces. The reason why reflection separation cannot significantly benefit from semantic guidance is that semantic information is not accurate enough when it is extracted from the image contaminated directly.

\textit{MTL scheme.} 
To verify the contribution of MTL, we use a Basic MTL scheme to jointly optimize semantic segmentation and reflection separation and make these two tasks share the same encoder, which is illustrated in Fig.~\ref{fig:MTL_model_choose} (c).
This effective operation can promote the performance of reflection separation (background/reflection) improved by $3.33/1.93$ PSNR and $0.038/0.116$ SSIM. \revrm{The visual comparison is illustrated in Fig~\ref{fig:ablation}-top (d) showing that the background layer and reflection layer are separated effectively.} However, it can be found that the arm of the left girl is mistakenly separated into the reflection layer. As a result, there are many residuals of background on the reflection layer. In the meantime, some parts with semantic meaning of the background are missing.

The performance of reflection separation is further improved by merging semantic guidance with MTL, as illustrated in Fig.~\ref{fig:MTL_model_choose} (d), which is our final model. Fig~\ref{fig:ablation}-top (f) shows clearer background at the region where semantic information guides. More specifically, persons in the background layer become clearer without artifacts on the arm \rev{(highlighted with dashed box)}. Meanwhile, the residuals are removed totally from the reflection layer. The PSNR of background/reflection is improved from $25.22/17.47$ to $26.02/18.51$ and SSIM is improved from $0.859/0.538$ to $0.875/0.592$. The semantic segmentation achieves better results by simultaneously learning to reflection removal. 

We further explore the relationship between semantic segmentation and different architectures. The quantitative comparison is reported in the 4-\textit{th} column of Table~\ref{tab:ablation_semantic_guid} and one group's results are illustrated in Fig.~\ref{fig:ablation}-top. 
Without multi-task learning, we find that the bottle in the input image cannot be recognized and separated correctly due to the reflection, which is shown in Fig.~\ref{fig:ablation}-top (d) and (f). 
\revrm{The MTL can improve the performance of semantic segmentation by 0.051 mIoU. The mIoU can be further improved from 0.546 to 0.584 by adding semantic guidance. }
\rev{It can be seen that the proposed MTL and semantic guidance not only can consistently improve the reflection removal performance on three datasets but also improve the performance of semantic segmentation (MTL: $\uparrow 0.151$ mIoU, Final version: $\uparrow 0.189$ mIoU). These quantitative results show the effectiveness of our proposed architecture.}

\begin{table}[htbp]
\begin{center}
	\setlength{\tabcolsep}{1.8mm}
	\caption{\rev{Segmentation performance of mIoU and reflection removal performance of SSIM and PSNR on the synthesized and real test set. MTL: multi-task learning. Guid.: semantic guidance.}}
	\label{tab:ablation_semantic_guid}
	\begin{tabular}{lcc|cccc|cc|cc}
		\toprule[1.3pt]
		& \multicolumn{ 2}{c|}{Architecture} & \multicolumn{ 4}{c|}{$\mathcal{D}_{syn}$} & \multicolumn{ 2}{c|}{\rev{$\mathcal{D}_{BKL}$}} & \multicolumn{ 2}{c}{\rev{$\mathcal{D}_{SIR^2}$}} \\
		Method & MTL & Guid.   & mIoU & \rev{Pixel acc.} &       SSIM &       PSNR &       \rev{SSIM} &       \rev{PSNR}    &   \rev{SSIM} &       \rev{PSNR} \\
		\hline
		\specialrule{0em}{1pt}{1pt}
		Baseline &              &              & 0.395 & \rev{75.33} & 0.821 & 21.67 & \rev{0.721} & \rev{18.79} & \rev{0.819} & \rev{21.68} \\
		\hline
		\specialrule{0em}{1pt}{1pt}
		SGR$^2$N & $\checkmark$ &			   & 0.546 & \rev{84.68} & 0.859 & 25.22 & \rev{0.801} & \rev{19.98} & \rev{0.877} & \rev{22.98} \\
		SGR$^2$N &  			& $\checkmark$ & 0.395 & \rev{75.33} & 0.833 & 24.70 & \rev{0.799} & \rev{21.02} & \rev{0.867} & \rev{22.62}\\
		SGR$^2$N & $\checkmark$ & $\checkmark$ & \textbf{0.584} & \rev{\textbf{88.79}} & \textbf{0.875} & \textbf{26.02} & \rev{\textbf{0.812}} & \rev{\textbf{22.46}} & \rev{\textbf{0.893}} & \rev{\textbf{23.81}}\\
		\bottomrule[1.3pt]
	\end{tabular}  
\end{center}
\end{table}

\textbf{Analysis of different loss functions.}
To analyze how each loss contributes to the final performance of our network on reflection separation, we conduct additional experiments on the feature loss Eqn.~\eqref{equ:loss_feat}, the pixel loss Eqn.~\eqref{equ:loss_pixel}, and adversarial loss Eqn.~\eqref{equ:loss_adv}. 
To make a detailed ablation study on feature loss $l_{feat}(\hat{\textbf{B}}, \textbf{B})$, we conduct ablation studies as follow:

\begin{enumerate}
	\item Totally remove the loss term. We remove $l_{feat}(\hat{R}, R)$ from the total loss, then $\hat{\textbf{R}}$ is generated through $\textbf{I} - \hat{\textbf{B}}$.
	\item Replace the loss term. We replace $l_{feat}(\hat{\textbf{B}}, \textbf{B})$ with the L1 difference between $\textbf{B}$ and $\hat{\textbf{B}}$.
\end{enumerate}

The $l_{feat}(\hat{\textbf{R}}, \textbf{R})$ is useful to the performance of reflection removal. To demonstrate this, we have made a similar ablation study on it. The quantitative results are reported in Table~\ref{tab:ablation_losses}. Note that the results of ablation studies on $l_{feat}(\hat{\textbf{R}}, \textbf{R})$ are marked with $\dag$. 

When we remove the feature loss $l_{feat}$, there are many residuals of reflections on the background layer, and some parts of the man's face are taken as reflection mistakenly, as shown in Fig.~\ref{fig:ablation}-bottom (b). The reason can be that $l_{feat}$ contains many perceptual constraints, which can constrain the low-level, mid-level, and high-level information of the background and reflection layers in the latent space \cite{ReflectionRemoval:Zhang2018PerceptualLosses}.  
When we remove the $l_{pix}$, some color-shifting occurs on the image, \eg, the man's shirt becomes darker, and the man's skin looks slightly over-saturated; the reason is that the $l_{pix}$ constrains the pixel-level appearance of the image.
The adversarial loss $l_{adv}$ further helps recover more natural backgrounds and reflections, as shown in the bottom of Fig.~\ref{fig:ablation} (d) and (e).

The quantitative comparisons are shown in Table.~\ref{tab:ablation_losses}. In the first row, we take the state-of-the-art method~\cite{ReflectionRemoval:Zhang2018PerceptualLosses} as reference. Removing different loss terms from the total loss, we test the SGR$^2$N under the same condition. In the second row, we remove feature loss and adversarial loss and take this model as a baseline. It can be found that the reference method~\cite{ReflectionRemoval:Zhang2018PerceptualLosses} performs better than our baseline. These results are reasonable because method \cite{ReflectionRemoval:Zhang2018PerceptualLosses} adopts various loss functions,~\viz~feature loss, adversarial loss, and exclusive loss.
Adding adversarial loss into the objectiveness, the performance of background/reflection improves $0.01/0.055$ SSIM and $0.42/1.25$ PSNR, and our method with these settings gets comparable results to the state-of-the-art. The 4-\textit{th} line of Table~\ref{tab:ablation_losses} shows that feature loss can improve $0.047/0.107$ SSIM and $3.48/1.81$ PSNR. 
\revrm{When (a) $l_{feat}(\hat{\textbf{R}}, \textbf{R})$ is removed, the SSIM/PSNR of the background layer drops from 0.875/26.02 to 0.861/25.58, and the SSIM/PSNR of the reflection layer drops from 0.592/18.51 to 0.450/16.93.} As for changing the $l_{feat}(\hat{\textbf{R}}, \textbf{R})$ to L1 loss, the performance is also decreased. For background layer, SSIM: 0.875$\to$0.871, PSNR: 26.02$\to$25.89. \revrm{For the reflection layer, SSIM: 0.592$\to$0.499, PSNR: 18.51$\to$17.87. }
The performance drops steadily without pixel loss. The combination of feature loss and adversarial loss can further boost the performance to $0.875/0.592$ SSIM and $26.02/18.51$ PSNR.
\rev{As shown in the last row, numerical results on other real benchmark datasets also show the combination of different loss terms achieves the best performance comprehensively. Visual results in a Fig.~\ref{fig:ablation} also show the clearer background and reflection layers, especially in the region with dashed boxes.}

\begin{table}[htbp]
\begin{center}
	\setlength{\tabcolsep}{2.5mm}
	\caption{\rev{Loss ablation experiments on the test set.}}
	\label{tab:ablation_losses}
	\begin{tabular}{lcc|cc|cc}
		\toprule[1.3pt]
		& \multicolumn{ 2}{c|}{$\mathcal{D}_{syn}$} & \multicolumn{ 2}{c|}{\rev{$\mathcal{D}_{BKL}$}} & \multicolumn{ 2}{c}{\rev{$\mathcal{D}_{SIR^2}$}} \\
		Method &      SSIM &       PSNR &       \rev{SSIM} &       \rev{PSNR}  & \rev{SSIM} & \rev{PSNR} \\
		\hline
		\specialrule{0em}{1pt}{1pt}
		Reference ~\cite{ReflectionRemoval:Zhang2018PerceptualLosses} & 0.835 & 22.66 & \rev{0.807} & \rev{22.37} & \rev{0.847} & \rev{22.82} \\
		\hline
		\specialrule{0em}{1pt}{1pt}
		Baseline 						   & 0.824 & 21.69 & \rev{0.801} & \rev{21.56} & \rev{0.849} & \rev{22.73} \\
		SGR$^2$N (w/o $l_{pix}$) 		   & 0.840 & 23.26 & \rev{0.791} & \rev{20.25} & \rev{0.811} & \rev{21.37}  \\
		SGR$^2$N (w/o $l_{feat}$ (a))	   & 0.834 & 22.93 & \rev{0.808} & \rev{22.37} & \rev{0.874} & \rev{23.24}\\
		SGR$^2$N (w/o $l_{feat}$ (b))	   & 0.860 & 24.99 & \rev{0.811} & \rev{22.46} & \rev{0.891} & \rev{23.80}\\
		SGR$^2$N (w/o $l_{feat}$ (a)$\dag$)& 0.861 & 25.58 & \rev{0.810} & \rev{22.41} & \rev{0.888} & \rev{23.61}\\
		SGR$^2$N (w/o $l_{feat}$ (b)$\dag$)& 0.871 & \underline{25.89} & \rev{\textbf{0.812}} & \rev{22.46} & \rev{\textbf{0.894}} & \rev{23.80}\\
		SGR$^2$N (w/o $l_{adv}$) 		   & \underline{0.872} & 25.17 & \rev{0.810} & \rev{\textbf{22.75}} & \rev{0.893} & \rev{\textbf{23.87}}\\
		SGR$^2$N & \textbf{0.875} & \textbf{26.02} & \rev{\textbf{0.812}} & \rev{\underline{22.46}} & \rev{\underline{0.893}} & \rev{\underline{23.81}}\\
		\bottomrule[1.3pt]
	\end{tabular}  
\end{center}
\end{table}

\subsection{Comparison with the State-of-the-Art Methods} \label{sec:cmp_sota}

In this section, we compare our SGR$^2$N against state-of-the-art methods including optimization-based approaches (LB14~\cite{ReflectionRemoval:Li2014Single}, AN17~\cite{ReflectionRemoval:Arvanitopoulos2017Single}, and YM19~\cite{ReflectionRemoval:Yang_2019_CVPR_Suppression}) and the learning-based methods (CEILNet~\cite{ReflectionRemoval:Fan2017A}, CRRN~\cite{ReflectionRemoval:Wan2018CRRN}, BDN~\cite{ReflectionRemoval:eccv18refrmv}, ZN18~\cite{ReflectionRemoval:Zhang2018PerceptualLosses}, ERRNet~\cite{ReflectionRemoval:Wei_2019_CVPR_ERRN}, WT19~\cite{ReflectionRemoval:Wen_2019_CVPR_Linear} and ICBLN~\cite{ReflectionRemoval:Li2020Cascaded}). For a fair comparison, we finetune these models on our training dataset if the training code has been published. We have also provided the numerical results of pretrained IBCLN (termed as IBCLN-p).

\textbf{Synthetic images.} 
Our synthetic test set contains 1969 \{input, background, and reflection\} triplets and corresponding semantic maps. We train all the existing methods on our synthetic training set if they were CNNs based and provided training code. Then we make quantitative and qualitative comparisons among these methods and ours.

\begin{figure*}[htbp]
	\begin{center}
		\includegraphics[width=\textwidth]{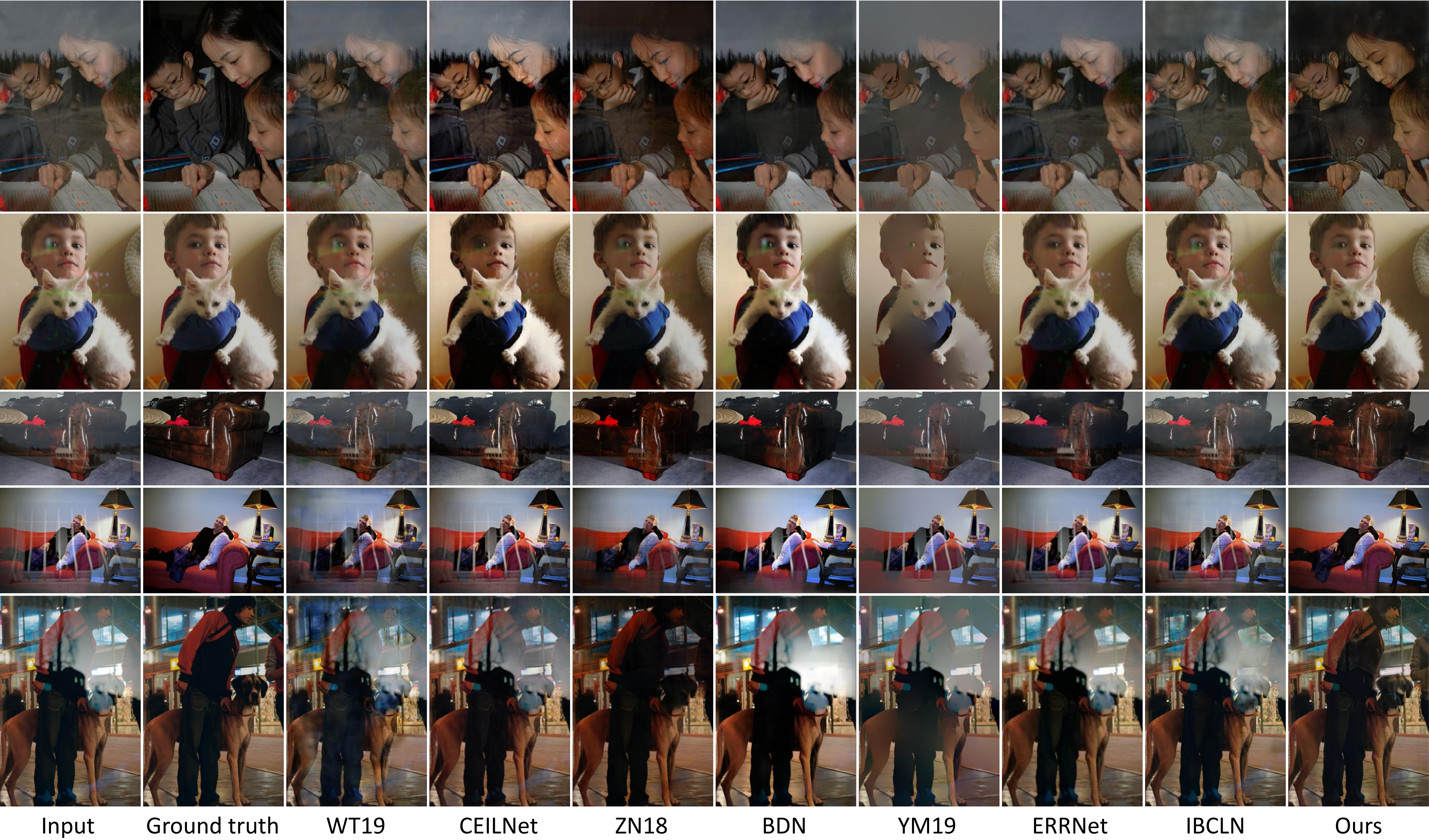}		
	\end{center}
	\caption{Visual background layers comparison of our method with four previous methods, evaluated on the synthetic dataset $\mathcal{D}_{syn}$. Our model generates superior results with cleaner local details.			
	}
	\label{fig:syn_images_cmp}
\end{figure*}

The quantitative results are shown in Table~\ref{tab:numerical_cmp_syn}.
The fine-tuned version of previous methods shows better numerical performance than the pre-trained version (Table~\ref{tab:numerical_cmp_syn} IBCLN-p v.s IBCLN).
We show stronger quantitative performance over previous works on both synthetic data, meanwhile, our method reaches a relatively fast runtime. 

Fig.~\ref{fig:syn_images_cmp} shows the qualitative results of different methods. It can be found that the state-of-the-art methods cannot remove reflection totally, while these reflections can be removed easily with semantic understanding via our SGR$^2$N. More specifically, WT19~\cite{ReflectionRemoval:Wen_2019_CVPR_Linear} cannot remove reflection on the area which has obvious semantic meaning. CEILNet~\cite{ReflectionRemoval:Fan2017A} can remove some reflections, however, it can be found other artifacts appear on the boy's face (the bottom of 5-\textit{th} column of Fig.~\ref{fig:syn_images_cmp}). ZN18~\cite{ReflectionRemoval:Zhang2018PerceptualLosses} can remove most reflections, but unpleasant color-shifting occurs on the girl's face. BDN~\cite{ReflectionRemoval:eccv18refrmv} enhances the color of the input image and smooths the reflection area, but the reflections aren't removed apparently. YM19~\cite{ReflectionRemoval:Yang_2019_CVPR_Suppression} over-smooths the input image and outputs unrealistic results. ERRNet~\cite{ReflectionRemoval:Wei_2019_CVPR_ERRN} can produce a relatively clean background layer, but there are still some reflection residuals on the output. To \rev{summarize}, our method produces the most compelling results with correct removal and cleaner local details. Furthermore, the previous methods show obvious quantitative performance margins below the proposed method on synthetic data. It can be explained that plenty of semantic objects greatly boosted the performance of reflection removal through our method.

\begin{table}[htbp]
	\begin{center}
		\setlength{\tabcolsep}{1mm}
		\caption{Quantitative comparison and runtime results among our method and other 8 prior works on $\mathcal{D}_{syn}$. Our model gets stronger quantitative performance over previous works and reaches relatively a fast runtime.}
		\label{tab:numerical_cmp_syn}
		\begin{tabular}{lcccccc}
			\toprule[1.3pt]
			& & \multicolumn{ 2}{c}{Background} & \multicolumn{ 2}{c}{Reflection} & \\
			Method & GPU &      SSIM &       PSNR &       SSIM &       PSNR & Runtime (\textit{s}) \\
			\hline
			\specialrule{0em}{1pt}{1pt}
			LM14~\cite{ReflectionRemoval:Li2014Single} &  & 0.779 & 19.19 & 0.3444 & 15.43 & 0.475 \\
			AN17~\cite{ReflectionRemoval:Arvanitopoulos2017Single} & & 0.781 & 19.23 & - & - & 99.38\\
			CEILNet~\cite{ReflectionRemoval:Fan2017A} & $\checkmark$ & 0.792 & 21.00 & - & - & 0.195\\
			BDN~\cite{ReflectionRemoval:eccv18refrmv} & $\checkmark$ & 0.824 & 19.67 & 0.345 & 11.29 & \textbf{0.024}\\
			ZN18~\cite{ReflectionRemoval:Zhang2018PerceptualLosses} & $\checkmark$ & 0.835 & 22.66 & 0.463 & 17.22 & 0.332\\
			YM19 ~\cite{ReflectionRemoval:Yang_2019_CVPR_Suppression} & & 0.796 & 20.35 & - & - & 0.270\\
			ERRNet ~\cite{ReflectionRemoval:Wei_2019_CVPR_ERRN} & $\checkmark$ & 0.827 & 22.71 & - & - & 0.719 \\
			WT19 ~\cite{ReflectionRemoval:Wen_2019_CVPR_Linear} & $\checkmark$ & 0.818 & 21.38 & - & - & 0.422\\
			IBCLN ~\cite{ReflectionRemoval:Li2020Cascaded} & $\checkmark$ & 0.817 & 22.09 & 0.407 & 13.40 & 0.189\\
			IBCLN-p~\cite{ReflectionRemoval:Li2020Cascaded} & $\checkmark$ & 0.808 & 21.88 & 0.371 & 13.29 & 0.189\\
			\hline
			\specialrule{0em}{1pt}{1pt}
			SGR$^2$N & $\checkmark$ & \textbf{0.878} & \textbf{26.02} & \textbf{0.592} & \textbf{18.51} & \underline{0.132} \\
			\bottomrule[1.3pt]
		\end{tabular}  
	\end{center}
\end{table}

\textbf{Real images.} 
To make a fair comparison, if the previous methods' pretrained models are published, we use the published model directly for comparison. Otherwise, We fine-tune these models on the real images under the same setting. 
The test set of Berkeley real image contains 20 \{input, background\} pairs, which are collected behind a portable glass~\cite{ReflectionRemoval:Zhang2018PerceptualLosses}.

\begin{figure*}[htbp]
	\begin{center}
		\includegraphics[width=\textwidth]{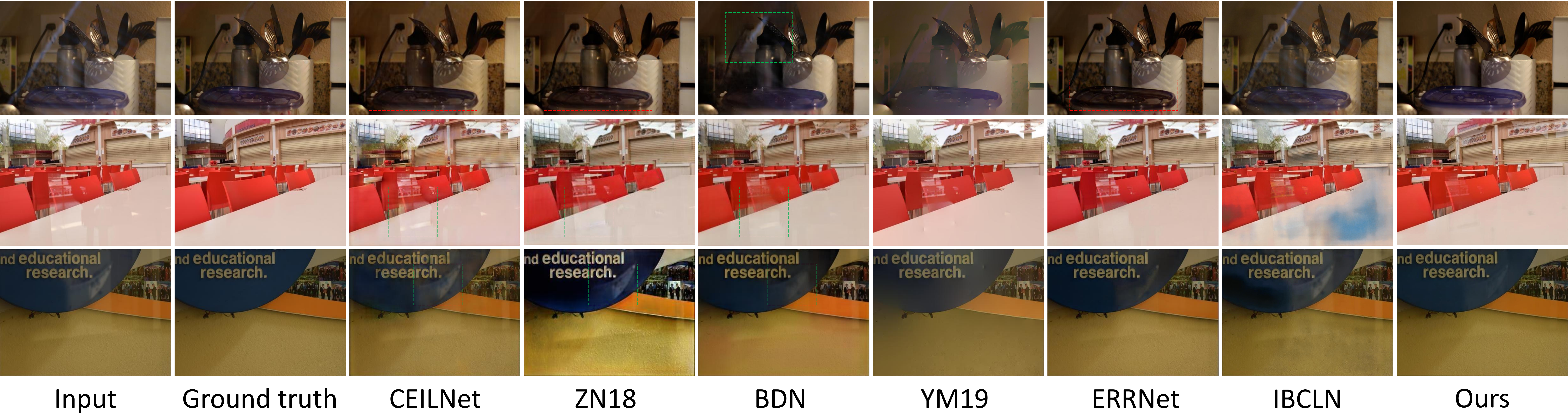}		
	\end{center}
	\caption{Visual background layers comparison of our method with 6 previous state-of-the-art methods, evaluated on the real images. 
	}
	\label{fig:real_images_cmp}
\end{figure*}

We further conduct comparisons on SIR$^2$~\cite{ReflectionRemoval:Wan2018CRRN}, which contains three sub-datasets. These three sub-datasets are captured under different conditions: (1) 20 controlled indoor scenes composed by solid objects; (2) 20 different controlled scenes on postcards; and (3) 55 wild scenes with ground truth. In the following, we compute the average PSNR/SSIM of these three sub-datasets for different methods.

Fig.~\ref{fig:real_images_cmp} shows the qualitative comparisons on the test set. Our method can handle images with complex semantic information. Furthermore, our method can generate cleaner background than other methods. 
The first row shows one visual comparison among state-of-the-art methods on Berkeley test set and the last two rows show the results on SIR$^2$ dataset. For the visual results on the first row, we observe that CEILNet~\cite{ReflectionRemoval:Fan2017A}, ZN18~\cite{ReflectionRemoval:Zhang2018PerceptualLosses}, and ERRNet~\cite{ReflectionRemoval:Wei_2019_CVPR_ERRN} can produce clean background. However, some details of color are missing, as illustrated in the red box. BDN~\cite{ReflectionRemoval:eccv18refrmv} and YM19~\cite{ReflectionRemoval:Yang_2019_CVPR_Suppression} can preserve the color detail of the input image, but BDN cannot remove reflection totally (as shown in the green boxes). YN19 produces over-smoothed images which are not realistic. Our method can overcome such limitations through semantic understanding and can reconstruct clean and correct background images. ICBLN introduces additional artifacts on the image, such as the white table suffered seriously in the second row.

Table \ref{tab:other_dataset_cmp_one_method} summarizes the results of all competing methods on these two real benchmarks \viz Berkeley test and SIR$^2$ benchmark. Note that Berkeley test images only provide the ground-truth for backgrounds; thus it is not adaptable to evaluate the performance on reflection images. \revrm{Although there are few semantic objects in the  $\mathcal{D}_{BKL}$, which leads to the results that our reflection removal module cannot get enough guidance from the semantic segmentation module, the SGR$^2$N still achieves comparable results.}
\rev{We find in this dataset, there are semantic categories \revii{that} are unseen in the training set. The semantic information cannot be used to guide reflection removal. In this situation, the performance of reflection removal decreases slightly. One possible way for this is to annotate more kinds of semantic labels. Even though, our SGR$^2$N still achieves comparable results.}
Meanwhile, our model shows superior performance on SIR$^2$ benchmark. Because the codes of Ma~\etal~\cite{ReflectionRemoval:Ma2019Jointly} and Kim~\etal~\cite{ReflectionRemoval:Kim2020PhysicallyBased} have not been published yet, we compare the numerical results (which are from their published paper) on the subset of the SIR$^2$ (\ie Wild Scenes). 
As reported in the SIR* in Table~\ref{tab:other_dataset_cmp_one_method}, our method shows consistently effective results. 

\begin{table}[htbp]
	\begin{center}
		\setlength{\tabcolsep}{2.6mm}
		\small
		\caption{Quantitative comparison results between the proposed method and other state-of-the-art methods on two real benchmarks. Note that best results are marked in bold, and the second places are underlined. Note that SIR$^2$* dataset denotes that only part of SIR$^2$ (\ie Wild scene) is used.}
		\label{tab:other_dataset_cmp_one_method}
		\begin{tabular}{clcccc}
			\toprule[1.3pt]
			
			& \multicolumn{ 1}{l}{} & \multicolumn{ 2}{c}{Background} & \multicolumn{ 2}{c}{Reflection} \\
			& \multicolumn{ 1}{l}{Method} &                              SSIM &       PSNR &          SSIM &       PSNR  \\
			\hline
			\specialrule{0em}{1pt}{1pt}
			
			\multirow{10}{*}{\rotatebox{90}{$\mathcal{D}_{BKL}$}}
			& \rev{Input (Reference)} & \rev{0.694} & \rev{17.66} & - & - \\
			& LM14~\cite{ReflectionRemoval:Li2014Single} & 0.563 & 17.54 & - & - \\
			& AN17~\cite{ReflectionRemoval:Arvanitopoulos2017Single} & 0.586 & 17.55 & - & - \\
			& CEILNet~\cite{ReflectionRemoval:Fan2017A} & 0.710 & 18.77 & - & - \\
			& BDN~\cite{ReflectionRemoval:eccv18refrmv} & 0.656  & 18.58 & - & - \\
			& ZN18~\cite{ReflectionRemoval:Zhang2018PerceptualLosses} & 0.807 & 22.37 & - & - \\
			& YM19 ~\cite{ReflectionRemoval:Yang_2019_CVPR_Suppression} & 0.640 & 18.05 & - & - \\
			& ERRNet ~\cite{ReflectionRemoval:Wei_2019_CVPR_ERRN} & \underline{0.811} & \textbf{22.84} & - & - \\
			& WT19 ~\cite{ReflectionRemoval:Wen_2019_CVPR_Linear} & 0.795 & 19.76 & - & - \\
			& IBCLN ~\cite{ReflectionRemoval:Li2020Cascaded} & 0.762 & 21.86 & - & - \\
			& \rev{RAGNet} ~\cite{ReflectionRemoval:li2020two} & \rev{0.793} & \rev{22.95} & \rev{-} & \rev{-} \\
			& SGR$^2$N & \textbf{0.812} & \underline{22.46} & - & - \\
			
			\hline
			\specialrule{0em}{1pt}{1pt}
			\multirow{11}{*}{\rotatebox{90}{$\mathcal{D}_{SIR{^2}}$}}
			& \rev{Input (Reference)} & \rev{0.801} & \rev{20.12} & \rev{0.410} & \rev{8.72} \\
			& LM14~\cite{ReflectionRemoval:Li2014Single} & 0.718 & 17.01 & 0.455 & 15.21 \\
			& AN17~\cite{ReflectionRemoval:Arvanitopoulos2017Single} & 0.716 & 18.96 & - & - \\
			& CEILNet~\cite{ReflectionRemoval:Fan2017A} & 0.819 & 21.65 & - & - \\
			& BDN~\cite{ReflectionRemoval:eccv18refrmv} & 0.842 & 21.76 & 0.270 & 8.77 \\
			& ZN18~\cite{ReflectionRemoval:Zhang2018PerceptualLosses} & 0.847 & 22.82 & 0.403 & 18.52 \\
			& YM19 ~\cite{ReflectionRemoval:Yang_2019_CVPR_Suppression} & 0.831 & 21.32 & - & - \\
			& \rev{CRRN ~\cite{ReflectionRemoval:Wan2018CRRN}} & \rev{0.884} & \rev{22.86} & - & - \\
			& ERRNet ~\cite{ReflectionRemoval:Wei_2019_CVPR_ERRN} & 0.890 & 23.59 & - & - \\
			& WT19 ~\cite{ReflectionRemoval:Wen_2019_CVPR_Linear} & 0.827 & 21.50 & - & - \\
			& IBCLN ~\cite{ReflectionRemoval:Li2020Cascaded} & 0.885 & 23.53 & 0.392 & 14.89 \\
			& IBCLN ~\cite{ReflectionRemoval:Li2020Cascaded}-p & 0.884 & 23.56 & 0.338 & 14.57 \\
			& \rev{SILS} ~\cite{ReflectionRemoval:liu2020separate} & \rev{0.821} & \rev{20.47} & \rev{0.425} & \rev{19.31} \\
			& \rev{RAGNet} ~\cite{ReflectionRemoval:li2020two} & \rev{0.890} & \rev{23.93} & \rev{0.511} & \rev{17.65} \\
			& SGR$^2$N & \textbf{0.893} & \textbf{23.81} & \textbf{0.607} & \textbf{19.54} \\
			
			\hline
			\specialrule{0em}{1pt}{1pt}
			\multirow{5}{*}{\rotatebox{90}{$\mathcal{D}_{SIR{^2}}$*}}
			& \rev{Input (Reference)} & \rev{0.868} & \rev{22.73} & \rev{0.441} & \rev{10.15} \\
			& Ma~\etal~\cite{ReflectionRemoval:Ma2019Jointly} & 0.903 & 24.48 & - & -  \\
			& Kim~\etal~\cite{ReflectionRemoval:Kim2020PhysicallyBased} & \textbf{0.905} & \underline{25.55} & - & -  \\
			& IBCLN ~\cite{ReflectionRemoval:Li2020Cascaded}  & 0.886 & 24.70 & 0.322 & 13.40 \\
			& IBCLN ~\cite{ReflectionRemoval:Li2020Cascaded}-p& 0.886 & 24.71 & 0.320 & 13.31 \\
			& \rev{SILS} ~\cite{ReflectionRemoval:liu2020separate} & \rev{0.833} & \rev{21.01} & \rev{0.444} & \rev{14.35} \\
			& \rev{RAGNet} ~\cite{ReflectionRemoval:li2020two} & \rev{0.880} & \rev{25.52} & \rev{0.492} & \rev{15.15} \\
			& SGR$^2$N & \textbf{0.905} & \textbf{25.93} & \textbf{0.544} & \textbf{17.10} \\
			
			\bottomrule[1.3pt]
		\end{tabular}  
	\end{center}
\end{table}

\begin{figure*}[htbp]
	\begin{center}
		\includegraphics[width=\linewidth]{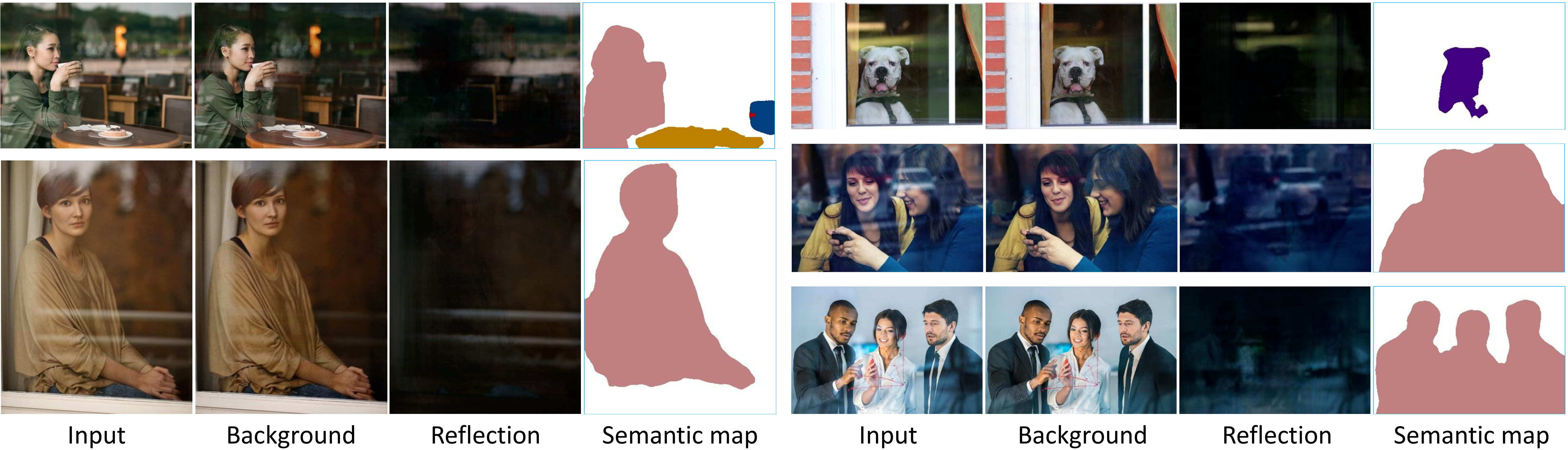}
	\end{center}
	\caption{More results on real images. The input images are from CEILNet~\cite{ReflectionRemoval:Fan2017A}.
	}
	\label{fig:more_results}
\end{figure*}

\begin{figure*}[htbp]
	\begin{center}
		\includegraphics[width=\linewidth]{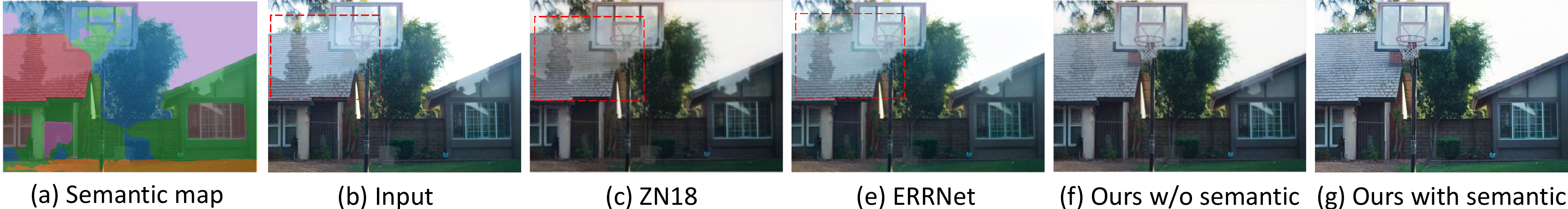}
	\end{center}
	\caption{
		Comparisons on scenes with complex semantics.
		Our method gets better visual quality than the completed methods. The performance is further improved with semantic guidance.
	}
	\label{fig:complex_semantic_ablation}
\end{figure*}

Our method also performs well for images with complex semantics. As illustrated in Fig.~\ref{fig:more_results} and Fig.~\ref{fig:complex_semantic_ablation}, our method can handle images with complex semantic information. It can be found that \rev{even if} there are more than 6 kinds of semantic objects in the input, our method can still perform well in this case. The semantic map is generated through DETR~\cite{Semantic:carion2020end} for reference. 

Fig.~\ref{fig:more_results} illustrates more visual results on real images, which are provided by~\cite{ReflectionRemoval:Fan2017A}. It can be observed that the proposed method generates clean backgrounds and clear reflections. Meanwhile, the semantic map is predicted correctly. These results show that our method can generalize well on various real scenes across different datasets.

\rev{\textbf{Alleviate the intensity sensitivity of SSIM and PSNR.}}
We notice in the qualitative comparison that the resulting images given by some comparing approaches are slightly darker than the ground truth. It may affect the \rev{SSIM and} PSNR result. 
Therefore, we align the lightness-histogram of all resulting images to be the same as the input such that the numerical comparison will not be biased by such as the brightness of the images. 

\begin{figure}[htbp]
	\begin{center}
		\includegraphics[width=0.6\linewidth]{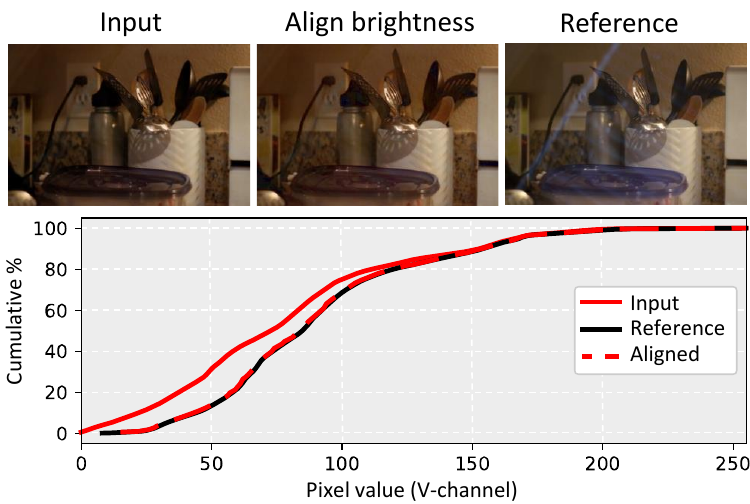}
	\end{center}
	\caption{
		Visual results of lightness-histogram alignment. The top row shows the visual comparison among the source image, the brightness matched image, and the reference image. The bottom row shows the cumulative result vs pixel value curve. It is observed that the brightness histogram of the matched image is closely aligned to the reference image.
	}
	\label{fig:histogram_alignment}
\end{figure}

To handle this, we have conducted an additional experiment as follows:

1) \textit{Histogram alignment.} We align the lightness-histogram of all resulting images to be the same as the input image. In detail, we first convert the resulting image and input from RGB space to HSV space. Since the V channel is highly related to the brightness, we align the V channel of the resulting image to the V channel of the input image. In the next, the original V channel of the resulting image is replaced with the result from the last step. Finally, we convert the modified resulting image back to RGB space. The visual results and alignment curves are illustrated in Figure~\ref{fig:histogram_alignment}. After histogram alignment, the resulting image becomes brighter.

\begin{figure}[htbp]
	\begin{center}
		\includegraphics[width=0.8\linewidth]{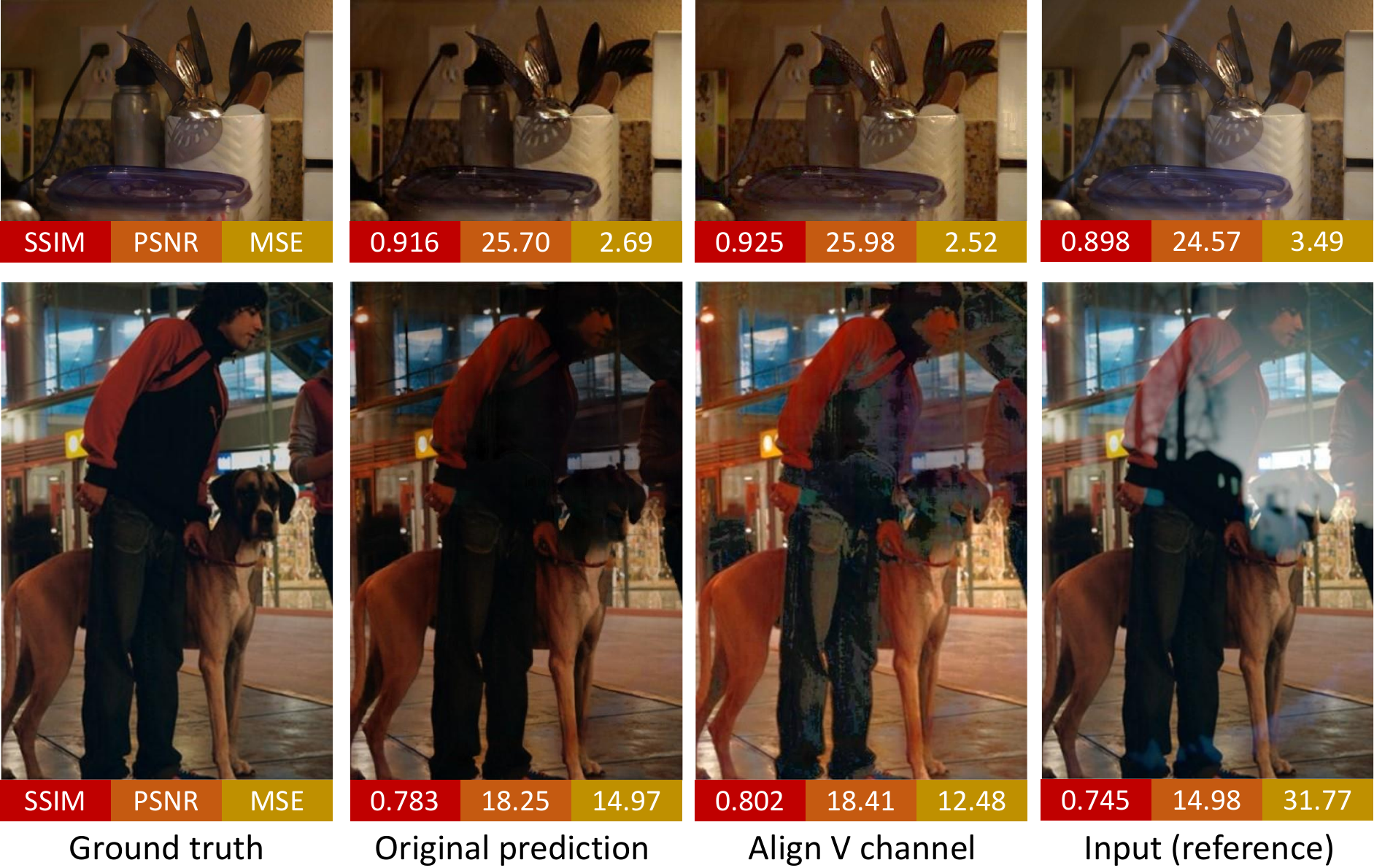}
	\end{center}
	\caption{
		Comparison of before/after brightness-histogram alignment. SSIM, PSNR, and MSE ($\times 1e3$) are used to measure the similarities between the predicted image and ground truth.
	}
	\label{fig:alignment_measure}
\end{figure}

2) \textit{Performance test.} We then computed the SSIM and PSNR between the aligned image and the corresponding ground truth. The alignment results will affect the PSNR and SSIM, visual results are shown in Figure ~\ref{fig:alignment_measure}. Quantitative results are shown in Table~\ref{tab:numerical_cmp}. As reported, the performances of all methods are slightly boosted in the synthetic dataset. Our method consistently generates better results over different datasets and layers.

In summary, after brightness alignment, our method still outperforms other competed methods, especially in $\mathcal{D}_{BKL}$.

\begin{table*}[htbp]
	\begin{center}
		\setlength{\tabcolsep}{2mm}
		\caption{Quantitative comparison results of background layer among our method and other 8 prior works, with and without brightness alignment to the input image. Our model gets consistently stronger quantitative performance over previous works.}
		\label{tab:numerical_cmp}
		\begin{tabular}{lcccccccc}
			\toprule[1.3pt]
			& \multicolumn{ 4}{c}{Synthetic dataset} & \multicolumn{ 4}{c}{Real dataset (Berkeley)} \\
			& \multicolumn{ 2}{c}{Original} & \multicolumn{ 2}{c}{Align brightness} & \multicolumn{ 2}{c}{Original} & \multicolumn{ 2}{c}{Align brightness} \\
			Method  &  SSIM &  PSNR &  SSIM &  PSNR &  SSIM &  PSNR &  SSIM &  PSNR \\
			\hline
			\specialrule{0em}{1pt}{1pt}
			LM14 	& 0.779 & 19.19 & 0.788 & 19.67 & 0.563 & 17.54 & 0.544 & 16.67 \\
			AN17 	& 0.781 & 19.23 & 0.792 & 19.59 & 0.586 & 17.55 & 0.541 & 17.02 \\
			CEILNet & 0.792 & 21.00 & 0.819 & 21.90 & 0.710 & 18.77 & 0.778 & 22.09 \\
			BDN 	& 0.824 & 19.67 & 0.827 & 20.08 & 0.656 & 18.58 & 0.670 & 19.21 \\
			ZN18 	& 0.835 & 22.66 & 0.844 & 24.35 & 0.807 & 22.37 & 0.819 & 22.80 \\
			YM19 	& 0.796 & 20.35 & 0.799 & 21.80 & 0.640 & 18.05 & 0.657 & 19.11 \\
			ERRNet 	& 0.827 & 22.71 & 0.838 & 23.79 & 0.811 & \textbf{22.84} & 0.825 & 22.90 \\
			WT19 	& 0.818 & 21.38 & 0.823 & 22.01 & 0.795 & 19.76 & 0.787 & 19.74 \\
			IBCLN 	& 0.817 & 22.09 & 0.819 & 23.17 & 0.762 & 21.86 & 0.780 & 22.04 \\
			\hline
			\specialrule{0em}{1pt}{1pt}
			SGR$^2$N & \textbf{0.878} & \textbf{26.02} & \textbf{0.880} & \textbf{26.99} & \textbf{0.812} & 22.46 & \textbf{0.830} & \textbf{22.99} \\
			\bottomrule[1.3pt]
		\end{tabular}  
	\end{center}
\end{table*}

\begin{figure}[htbp]
	\begin{center}
		\includegraphics[width=0.6\linewidth]{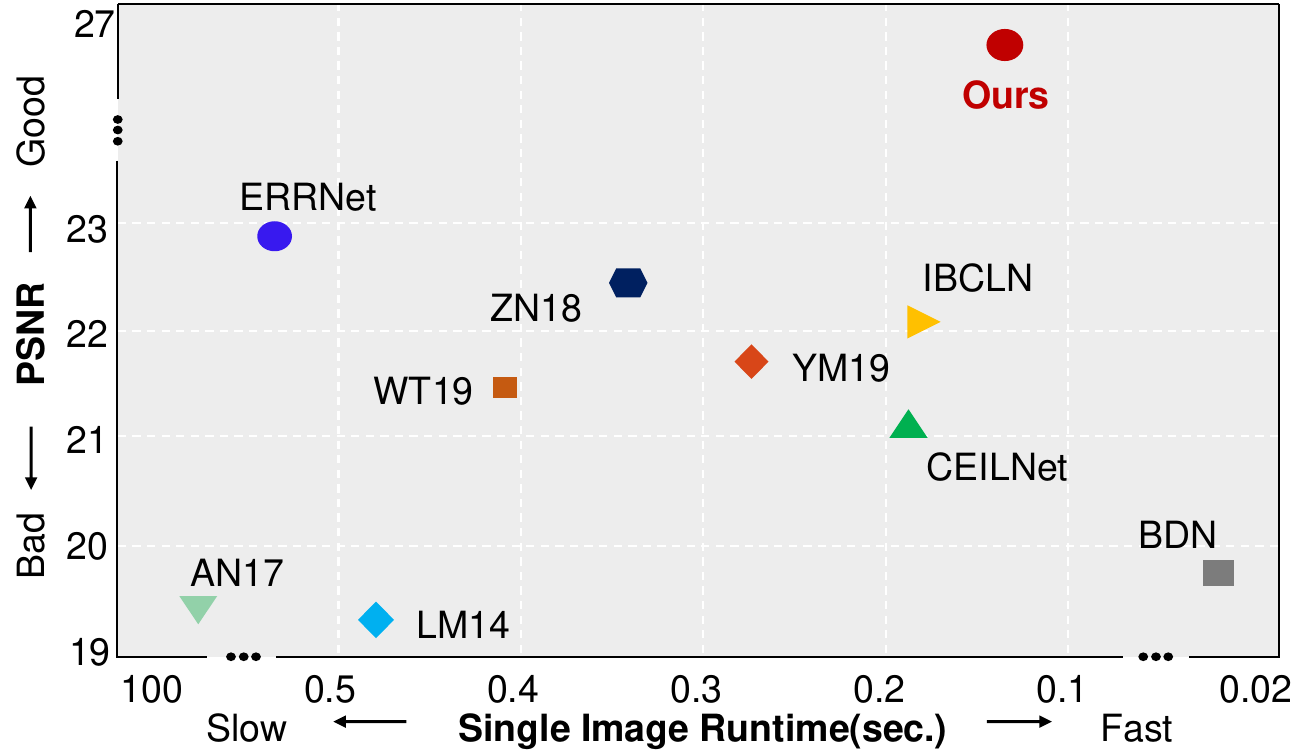}	
	\end{center}
	\caption{Comparison with previous methods: PSNR versus single image runtime. Numbers are taken from Table~\ref{tab:numerical_cmp_syn}. 
		Our method outperformed previous methods in a comprehensive way.}
	\label{fig:psnr_vs_time}
\end{figure}

\textbf{Inference time.} 
We further test the running time of prior works and ours and present the result in the last column of Table~\ref{tab:numerical_cmp_syn}. 
We test different approaches on $\mathtt{Ubuntu 18.04}$, with an Intel \textregistered  $\text{Core}^{TM}$ i7-7700 CPU and a $\mathtt{GeForce GTX 1080}$ GPU card. 
The comprehensive comparison is illustrated in Fig.~\ref{fig:psnr_vs_time}.
In detail, methods LM14~\cite{ReflectionRemoval:Li2014Single}, AN17~\cite{ReflectionRemoval:Arvanitopoulos2017Single}. and YM19~\cite{ReflectionRemoval:Yang_2019_CVPR_Suppression} are tested on CPU with Mathlab 2017b~\cite{API:MATLAB}. 
The other methods are tested on GPU:
CEILNet~\cite{ReflectionRemoval:Fan2017A} is implemented through Torch~\cite{API:BigLearn2011Torch7}.
YG18~\cite{ReflectionRemoval:eccv18refrmv}, BDN~\cite{ReflectionRemoval:eccv18refrmv}, ERRNet~\cite{ReflectionRemoval:Wei_2019_CVPR_ERRN}, and WT19~\cite{ReflectionRemoval:Wen_2019_CVPR_Linear} are implemented via PyTorch~\cite{API:paszke2017automatic}.
Method ZN18~\cite{ReflectionRemoval:Zhang2018PerceptualLosses} is implemented through Tensorflow.
We find our method decreased the computational cost, in terms of parameters and inference time. To demonstrate this, we have conducted additional comparisons between our methods and the other two state-of-the-art methods~\cite{ReflectionRemoval:Zhang2018PerceptualLosses,ReflectionRemoval:Wei_2019_CVPR_ERRN}. Detailed results are listed in Table~\ref{tab:architure_param} \rev{supplemental material}.

\subsection{Relationship Between the Semantic Segmentation and Reflection Removal}

To understand how the quality of semantic information affects reflection removal, we conduct an additional experiment to more deeply explore the relationship between semantic guidance and reflection removal. 
The experiment is organized as follows:
\begin{enumerate}
	\item Reflection removal with \textit{accurate} semantics as guidance. We use the features from the semantic estimation branch of our final model. In this case, semantic features are correctly computed.
	\item Reflection removal \textit{without} semantic guidance. In this case, we remove the semantic features directly. 
	\item Reflection removal with \textit{wrong} semantics as guidance. To demonstrate the effect of wrong semantics, we pre-compute the semantic features of another image. Then we replace the input of the semantic guidance module with these features. In this way, wrong semantics are taken as guidance for reflection removal.   
\end{enumerate}

\begin{figure}[htbp]
	\begin{center}
		\includegraphics[width=0.8\linewidth]{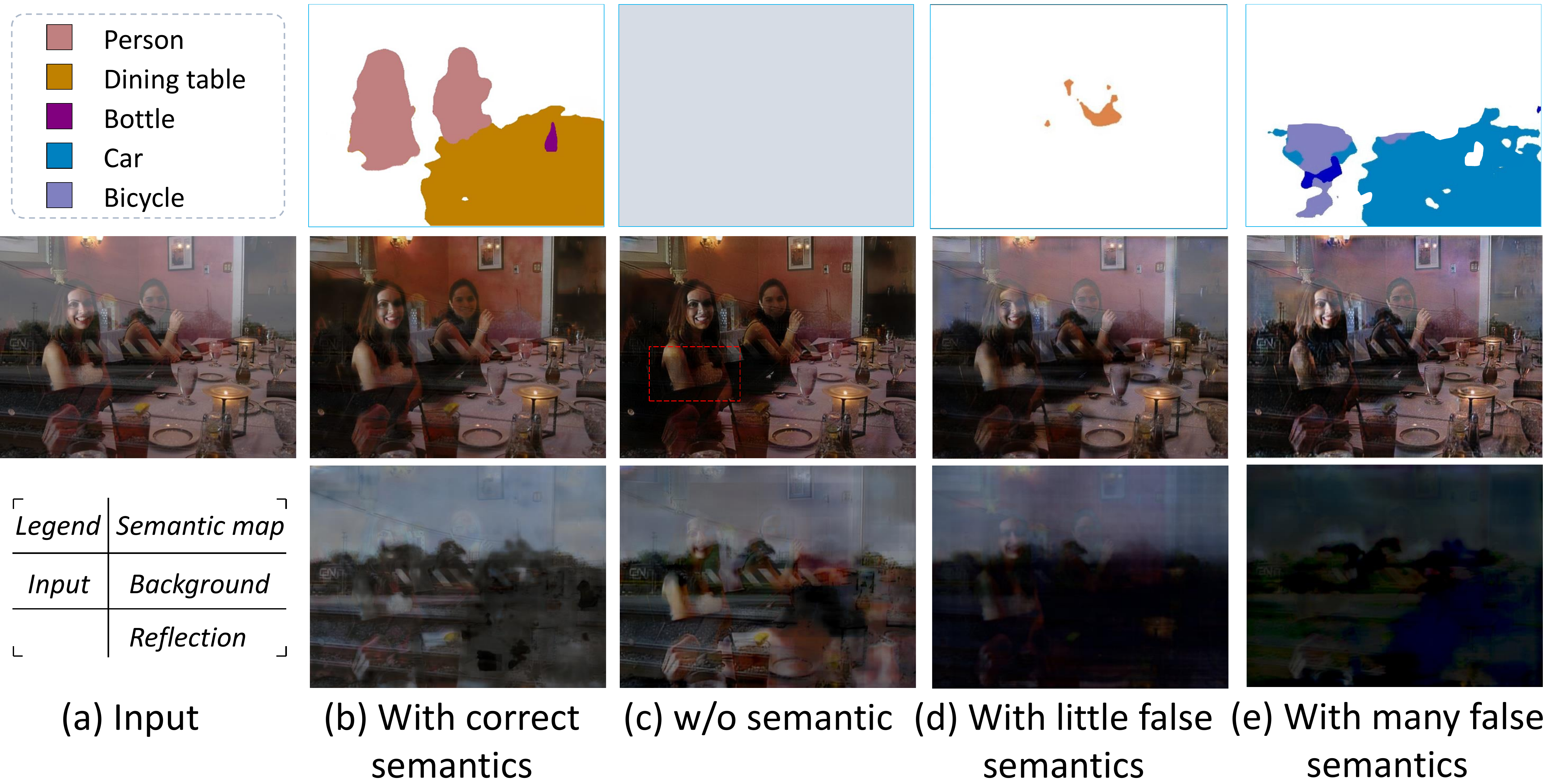}
	\end{center}
	\vspace{-0.3cm}
	\caption{
		The effect of semantic guidance quality in final reflection removal results. (b) with correct semantic guidance, (c) without semantic guidance, (d)-(e) with wrong semantic guidance.
	}
	\label{fig:semantics_analysis}
\end{figure}

The visual results are illustrated in Fig.~\ref{fig:semantics_analysis}. Then one can find the results that 
\begin{enumerate}
	\item When accurate semantics are used as guidance, which is illustrated in Fig.~\ref{fig:semantics_analysis} (b), reflection can be well removed and the background tends to be clean and of high quality.
	\item When the semantic information is removed, as shown in Fig.~\ref{fig:semantics_analysis} (c), the reflection can be still removed from the mixed image. But there may leave few contextual residuals of reflection, as illustrated in the dashed box.
	\item When the wrong semantic features are used to guide reflection removal, as shown in Fig.~\ref{fig:semantics_analysis} (d)-(e), the results suffer some artifacts.
\end{enumerate}
In summary, semantic information is important to reflection removal, and different semantics will affect the reflection separation performance. 
In addition, the quality and accuracy of the semantic information are jointly optimized with our method. As shown in Fig.~\ref{fig:alpha_vs_segmentation}, the semantic estimation is relatively robust in most cases.

\subsection{Exploration of Performance v.s the Reflection Strength}

In this section, we make experiments on the relationship between the SGR$^2$N performance and the strength level of the reflection layer.
We generate a series of image quadruples of $\{\mathbf{I}_t, \mathbf{B}_t, \mathbf{R}_t, \mathbf{S}_t\}^N_{t=1}$, where $\mathbf{I} = (1-\alpha)\mathbf{B} + \alpha \mathbf{R}$, \revrm{$\alpha=0.1, 0.2, \dots 0.9$,} \rev{where $\alpha$ is \revii{a} reflection strength-relevant parameter. We set $\alpha=0.1, 0.2, \dots, 0.9$ in this experiment to generate 9 groups of datasets, respectively. 
Here images are randomly selected from Pascal VOC~\cite{Semantic:Everingham2010pascalVOC}.
Each group contains 1000 quadruples (\textbf{I}, \textbf{B}, \textbf{R}, and \textbf{S}) for training and 200 quadruples for test. For each group, we train and test our method under the same condition.
Then we compare the reflection removal and semantic performance (pixel accuracy and mIoU) v.s $\alpha$ through different settings.} 
We compare the final mIoU of DeeplabV3+~\cite{Semantic:Chen2018DeepLabV3plus} \rev{(pretraiend on PascalVOC dataset)} and the final PSNR of our baseline~\cite{ReflectionRemoval:Zhang2018PerceptualLosses} on such images.

\begin{figure}[htbp]
	\centering
	\begin{subfigure}{.495\textwidth}
		\includegraphics[width=\textwidth]{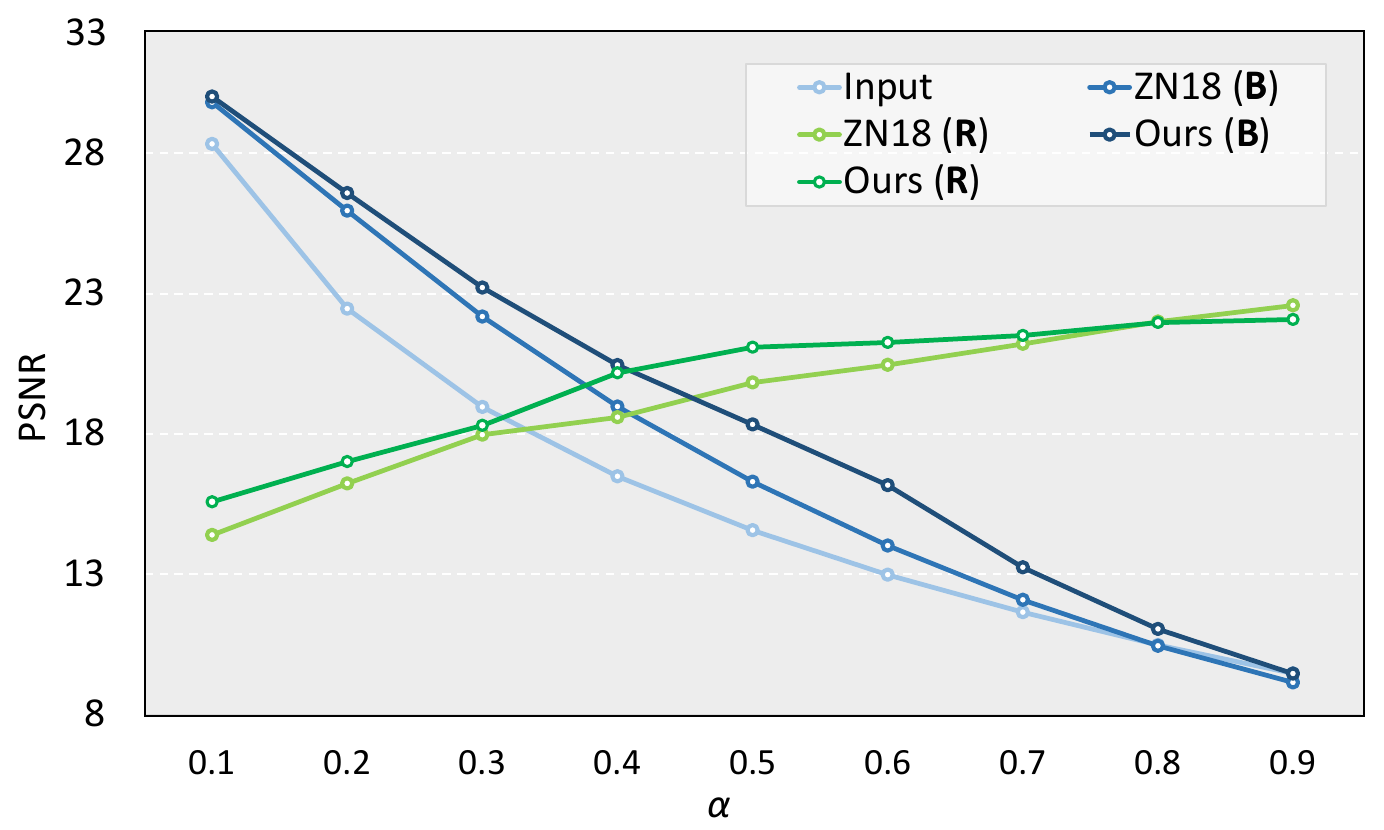}
		\caption{Reflection removal results (PSNR) in images with different reflection intensities.}
		\label{fig:alpha_with_sepatation}
	\end{subfigure}
	\hfill
	\begin{subfigure}{.495\textwidth}
		\includegraphics[width=\textwidth]{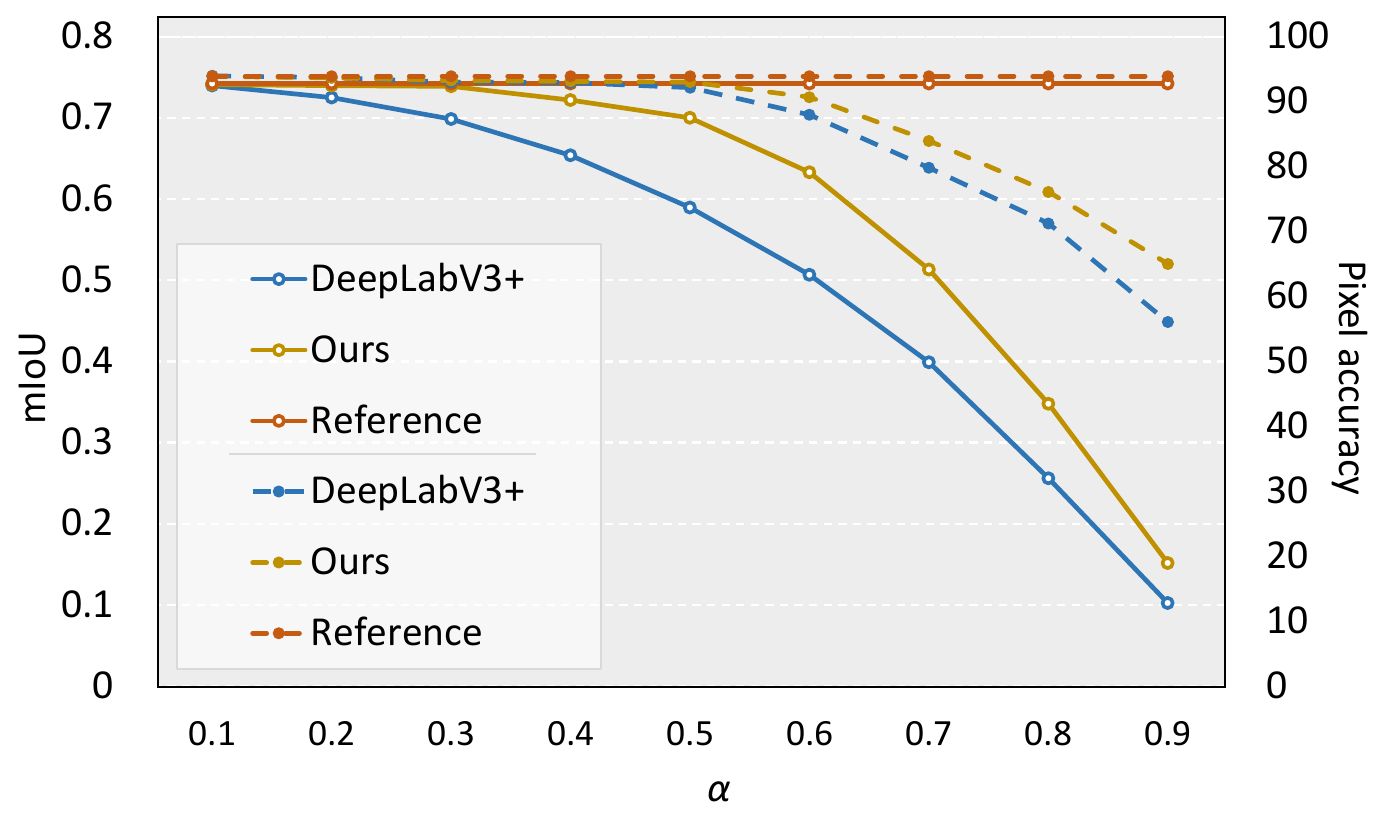}
		\caption{Semantic segmentation results (mIoU and \rev{pixel accuracy}) in images with different reflection intensities.}
		\label{fig:alpha_vs_segmentation}
	\end{subfigure}
	\caption{Results of reflection removal (a) and semantic segmentation (b) v.s different reflection strengths.}
\end{figure}

As presented in Fig.~\ref{fig:alpha_with_sepatation}, the proposed SGR$^2$N performs a higher score than the baseline in most cases with different $\alpha$ values.
Furthermore, the SGR$^2$N performs a more robust result to different intensities of reflection layer, as illustrated in Fig.~\ref{fig:alpha_vs_segmentation}. \rev{We find that reflection influences semantic accuracy slightly when reflection strength is lower than 0.5. Otherwise, the reflection will influence semantic accuracy. Note that such scenarios are also hard cases for reflection removal because the input image is dominated by reflection scenes.}

\subsection{Failure Cases and Discussion}

\begin{figure}[htbp]
	\begin{center}
		\includegraphics[width=\linewidth]{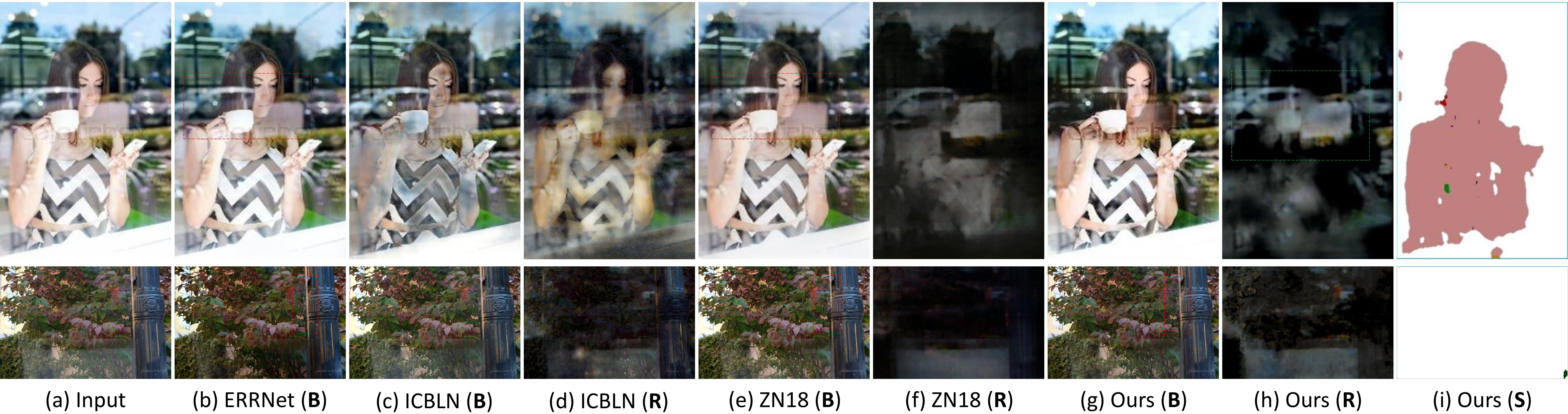}
	\end{center}
	
	\caption{Two typical failure cases for our method. (1) The reflection intensity is too strong. 
		(2) The input image lacks meaningful semantic information. Our method can still produce comparable and even superior results than the state-of-the-art methods ERRNet~\cite{ReflectionRemoval:Wei_2019_CVPR_ERRN}, IBCLN~\cite{ReflectionRemoval:Li2020Cascaded} and ZN18~\cite{ReflectionRemoval:Zhang2018PerceptualLosses}.
	}
	\label{fig:failure_case}
\end{figure}

Although the SGR$^2$N achieves the state-of-the-art on these three datasets, there are challenging cases illustrated in Fig.~\ref{fig:failure_case}.
One of the challenging scenarios where the reflection in the input is too strong, the background is contaminated heavily and our model may not separate layers successfully.
The cascade update scheme ICBLN \cite{ReflectionRemoval:Li2020Cascaded} can not separate the reflection correctly. As shown in the first row of Fig.~\ref{fig:failure_case}, the low-frequency part of the input image is incorrectly separated into reflection layer.
Note that reflections cannot be totally removed by these methods, but still, our result is superior to \cite{ReflectionRemoval:Zhang2018PerceptualLosses} (\eg, the person in the background is more distinguishable and the reflection layer is clearer to the other two state-of-the-art methods). 

\subsection{Extend Applications}
\begin{figure}[htbp]
	\begin{center}
		\includegraphics[width=\linewidth]{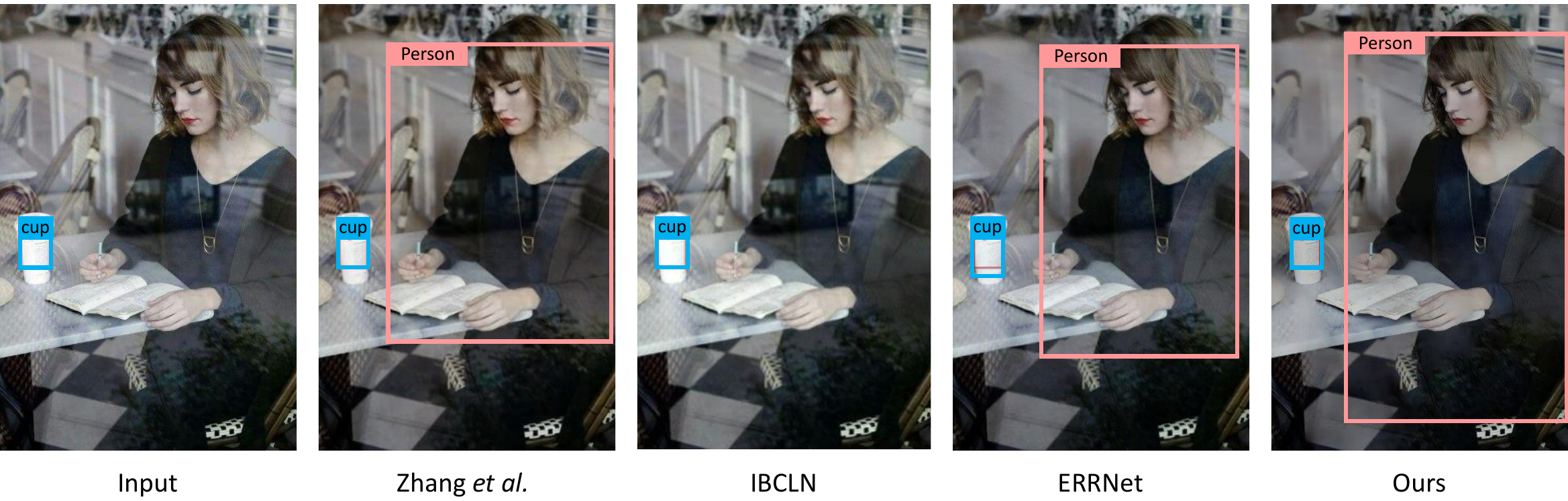}
	\end{center}
	\caption{We run YOLO-v3~\cite{ImageProcess:yolov3} on object detection to verify the effectiveness of SGR$^2$N. Left: the visual result of images with reflection; \rev{Right: the results of predicted background from the different methods.}}
	\label{fig:extend_app-detection}
\end{figure}

We extend the proposed method on object detection.
To verify how reflection removal benefits object detection, we conduct additional experiments by using the state-of-the-art objection detection approach YOLO-v3~\cite{ImageProcess:yolov3}. Specifically, we generate 500 groups of images with bounding box annotations based on Pascal VOC~\cite{Semantic:Everingham2010pascalVOC}.
We use mean average precision (mAP) over all 20 classes as evaluation metrics for objection detection.
Quantitative and qualitative comparisons are shown in Table.~\ref{tab:extend_app-detection} and Fig.~\ref{fig:extend_app-detection}, respectively. Table.~\ref{tab:extend_app-detection} shows that the mAP in reflection contaminated image drops over 12\%, and the results are greatly improved after reflection removal.
The detection results on reflection-removed output $\hat{\textbf{B}}$ hold higher numerical performance and better visual appearance. In detail, as shown in Fig.~\ref{fig:extend_app-detection}, the person can be predicted accurately when the reflection is removed. \rev{We find the detection results on our method are relatively accurate because our method produces more clean background than compared method.}

\begin{table}[htbp]
	\centering
	\small
	\setlength{\tabcolsep}{10pt}
	\caption{Pascal VOC object detection results. The mAP of the clear background is used as the baseline. }
	\label{tab:extend_app-detection}
	\begin{tabular}{lccc}
		\toprule[1.3pt]
		Input & \textbf{B} & \textbf{I} & $\hat{\textbf{B}}$         \\ \hline
		\specialrule{0em}{1pt}{1pt}
		mAP   & 47.17 & 34.75 ($\downarrow$ 12.42\%) & 42.83 ($\downarrow$ 4.34\%) \\ 
		\bottomrule[1.3pt]
	\end{tabular}
\end{table}

\section{Conclusion and Discussion}

In this paper, we have presented an approach to use semantic clues for the task of single image reflection separation. 
We first explore the relationship between semantic segmentation and reflection removal and applied high-level guidance explicitly.
We design a deep encoder-decoder network for image feature extraction and use a semantic segmentation network in parallel. Then with the two kinds of information fused together, our separation network can correctly separate the background layer and reflection layer.
We evaluate our method with other prior works extensively on three different datasets. The comparison result shows that our approach can outperform the existing methods both quantitatively and visually on all three datasets. Like other methods, the extreme cases where the reflection layer is too strong or the background layer contains texture-less objects are considered as the limitation of our method and we leave such cases as our future directions.

\bibliographystyle{ACM-Reference-Format}
\bibliography{sample-base}

\newpage
\appendix

{\centering{\textbf{\LARGE \rev{Online Appendix}}}}

\section{More Visual Results on Real Benchmark Datasets}

\begin{figure*}[b]
	\includegraphics[width=\linewidth]{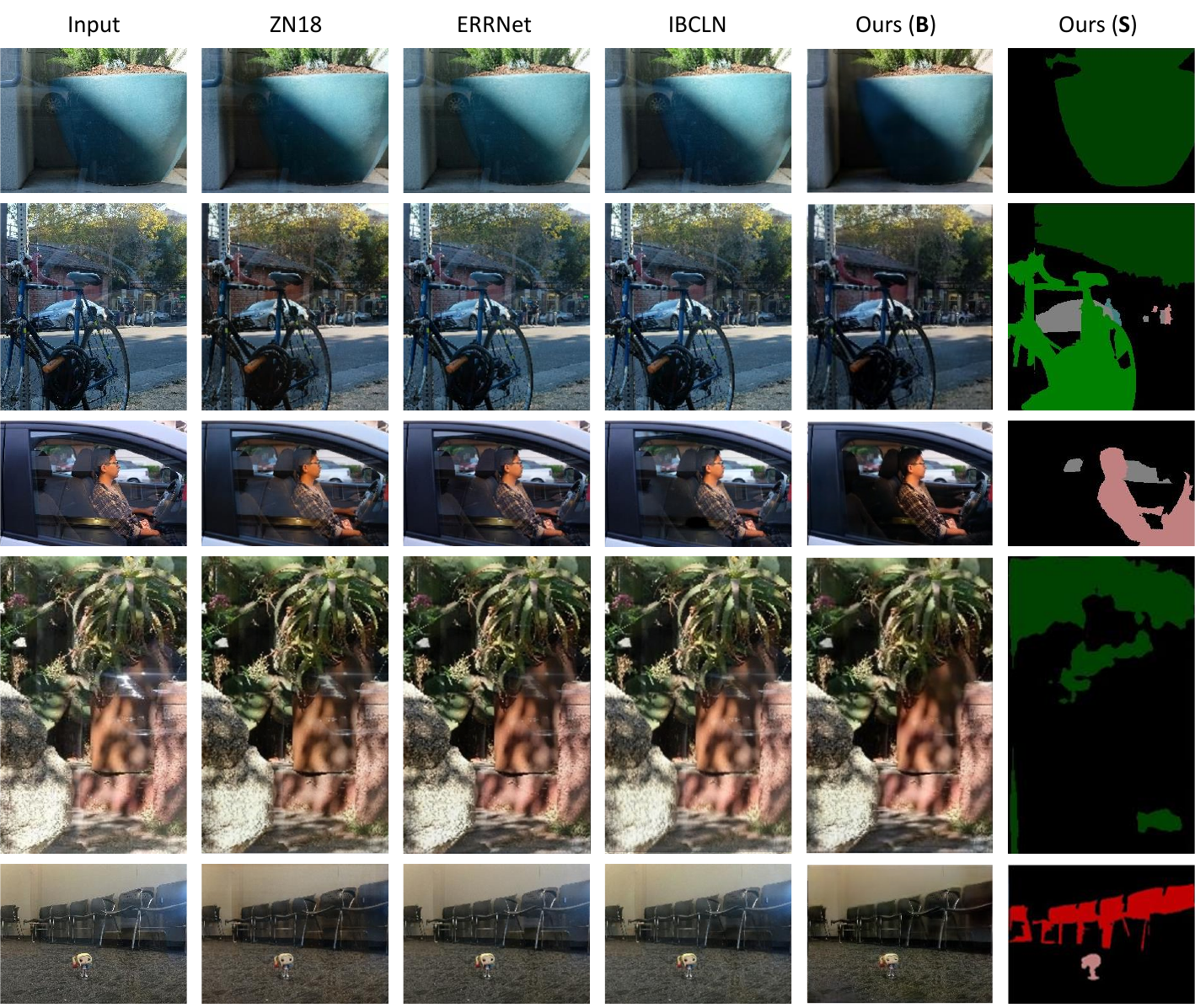}
	\caption{\rev{Qualitative comparison among state-of-the-art reflection removal methods and ours over $\mathcal{D}_{BKL}$. Please zoom in on the screen for the details. We have attached our semantic segmentation for reference.}}
	\label{fig:rr-result-bkl}
\end{figure*}

\begin{figure*}[htbp]
\includegraphics[width=\linewidth]{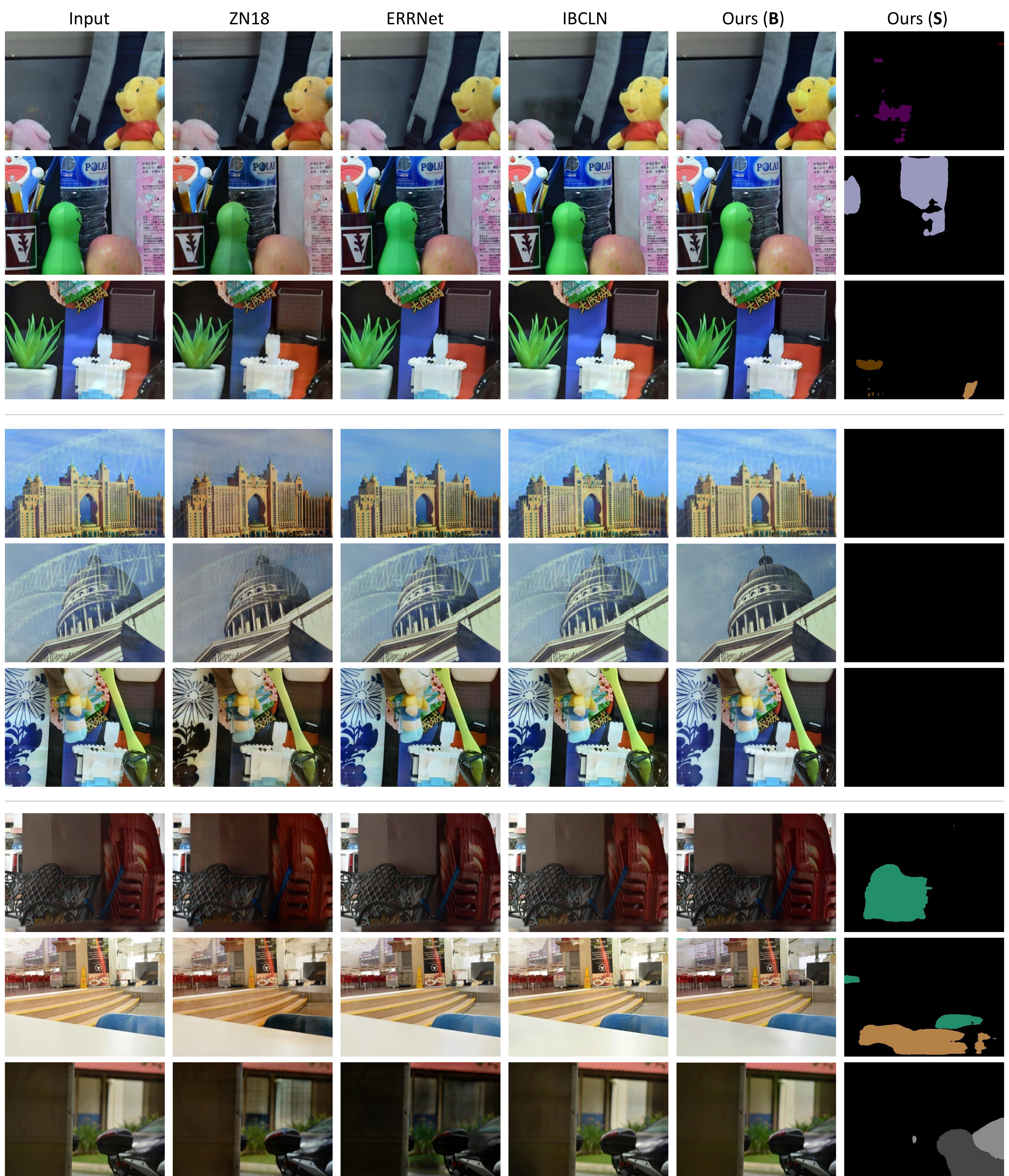}
\caption{\rev{Qualitative comparison among state-of-the-art reflection removal methods and ours over $\mathcal{D}_{SIR^2}$. We have attached our semantic segmentation for reference.}}
\label{fig:rr-result-sir2}
\end{figure*}

\rev{We show more visual results in Fig.~\ref{fig:rr-result-bkl} and Fig.~\ref{fig:rr-result-sir2} on real benchmark dataset, including $\mathcal{D}_{BKL}$ and $\mathcal{D}_{SIR^2}$. 
Various scenarios with reflections are presented. Here we compare three state-of-the-art reflection removal methods, ZN18~\cite{ReflectionRemoval:Zhang2018PerceptualLosses}, ERRNet~\cite{ReflectionRemoval:Wei_2019_CVPR_ERRN}, IBCLN~\cite{ReflectionRemoval:Li2020Cascaded}, \rev{SILS~\cite{ReflectionRemoval:liu2020separate}, and RAGNet~\cite{ReflectionRemoval:li2020two}}. 
In Fig.~\ref{fig:rr-result-bkl}, we find our method generates precise semantic segmentation (\textbf{S}). Also, compared with the other methods, our method yields favorable results. For instance, the reflections on the semantic objects, \eg, vase(first row), the chair (last row), \etc, are all removed properly by our method.}

\rev{
We have also presented the visual results on $\mathcal{D}_{SIR^2}$ in Fig.~\ref{fig:rr-result-sir2}. This figure contains three types of scenes. From top to down, they are solid object scenes, postcard scenes, and wild scenes. We find out method can generate comparable results when the semantic information is limited, especially in the postcards scenes. When there is some semantic information for guidance, our method produces more favorably visual results than other methods. These results show the flexibility and generalization ability of our framework and semantic guidance module.}

\section{More Details About the Architecture of Different Methods}
\begin{table*}[htbp]
	\begin{center}
		\setlength{\tabcolsep}{0.9mm}
		\small
		\caption{Number of model parameters and inference time comparison of different methods.}
		\label{tab:architure_param}
		\begin{tabular}{l cc cc cc cc}
			\toprule[1.3pt]
			\multirow{2}{*}{Method}  & \multicolumn{ 2}{c}{Encoder} & \multicolumn{ 2}{c}{Decoder} & \multicolumn{ 2}{c}{Discriminator} & Total & Inference\\
			& Params(M) & FLOPs(G) & Params(M) & FLOPs(G) & Params(M) & FLOPs(G) & Params(M) & Time (ms)\\
			\hline
			\specialrule{0em}{1pt}{1pt}
			\multirow{2}{*}{ZN18~\cite{ReflectionRemoval:Zhang2018PerceptualLosses}}  & \multicolumn{ 2}{c}{\mygray{VGG-19}} & \multicolumn{ 2}{c}{\mygray{FCN with dilations}}	& \multicolumn{ 2}{c}{\mygray{5-layer CNNs}} & \multirow{2}{*}{150.34} & \multirow{2}{*}{432} \\
			& 143.67     & 19.63  		 & 3.90			& 48.84						& 2.77 				& 		2.71     & & \\
			\multirow{2}{*}{ERRNet~\cite{ReflectionRemoval:Wei_2019_CVPR_ERRN}}& \multicolumn{ 2}{c}{\mygray{VGG-19}} & \multicolumn{ 2}{c}{\mygray{FCN with designed modules}} & \multicolumn{ 2}{c}{\mygray{12-layer CNNs}} & \multirow{2}{*}{172.55}  & \multirow{2}{*}{719} \\
			& 143.67 	& 19.77  		 &    18.95  	& 	337.40          		& 9.93 				& 		5.31    & & \\
			\multirow{2}{*}{Ours} & \multicolumn{ 2}{c}{\mygray{ResNet-101 + ASPP}} & \multicolumn{ 2}{c}{\mygray{Image + semantic decoder}} & \multicolumn{ 2}{c}{\mygray{3-layer CNNs}} & \multirow{2}{*}{\textbf{63.60}}  & \multirow{2}{*}{\textbf{132}} \\
			& \textbf{57.93}      &  \textbf{12.87}  		 &  4.46   & 41.12           	    & \textbf{1.21}     			&        1.77 & & \\
			\bottomrule[1.3pt]
		\end{tabular}  
	\end{center}
\end{table*}

As shown in Table \ref{tab:architure_param}, ZN18 and ERRNet use VGG-19 as encoder,
which uses much more parameters than that in our encoder, \ie, 143.67M and 57.93M. Besides, the sum of parameters in the decoder of ZN18 is 3.90M, and ours is 4.46M, whereas ERRNet uses 18.95M parameters in the decoder. 
Therefore, our method has fewer parameters in the total model. Furthermore, Our method is also inference faster \rev{than} these two state-of-the-art methods. More intuitive comparisons among different methods are illustrated in Fig.~\ref{fig:psnr_vs_time}.

\section{Extend Applications on Low-level Vision Tasks}

\begin{figure*}
	\includegraphics[width=0.85\linewidth]{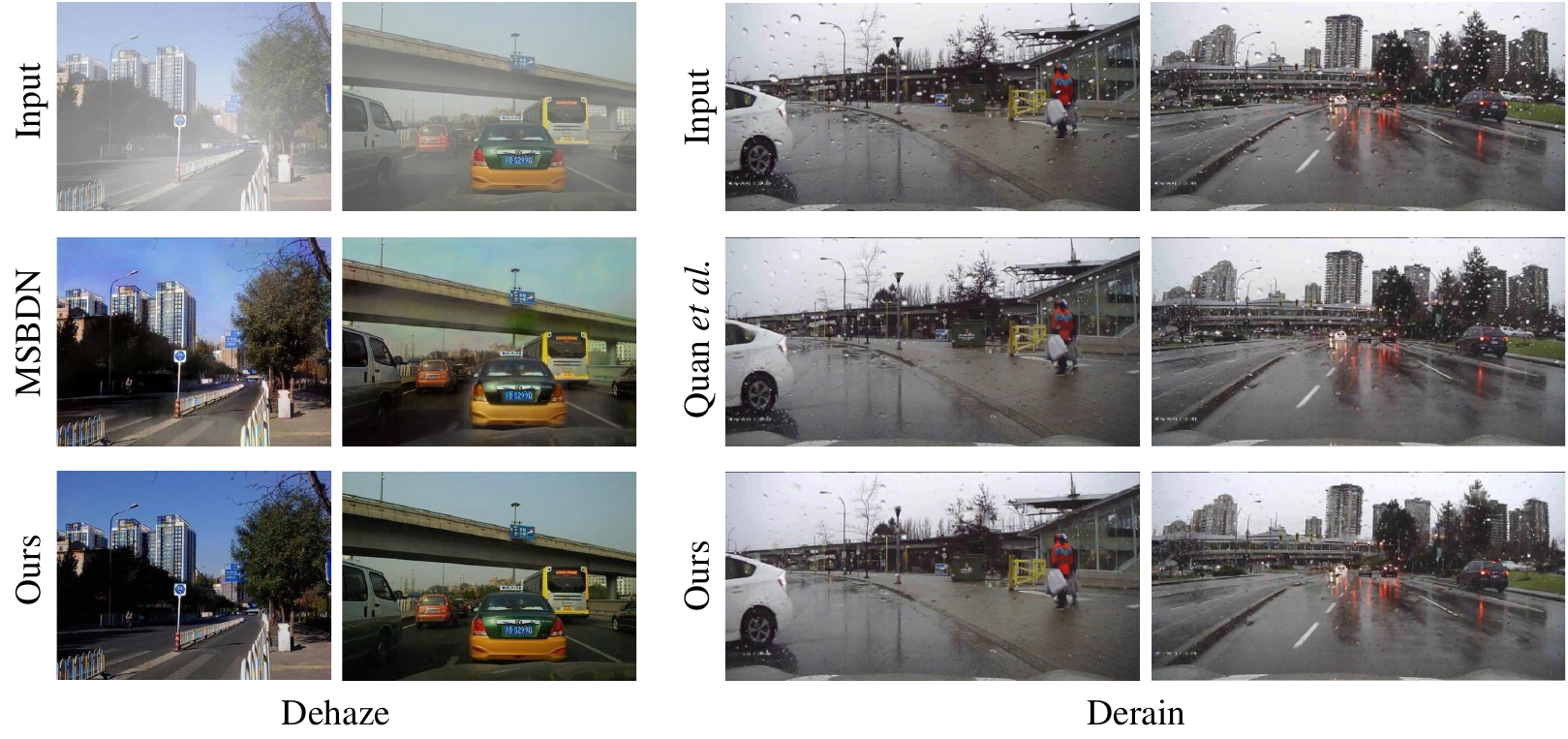}
	\caption{\rev{Generalization ability of our method on dehazing and deraining tasks.}}
	\label{fig:dehaze_derain}
\end{figure*}

\begin{table*}[htbp]
	\caption{\rev{Quantitative results of our method on dehazing and deraining tasks.}} 
	\label{tab:app_LLV}
	\begin{center}
		\small
		\setlength{\tabcolsep}{4mm}
		\begin{tabular}{l cc | c cc}
			\toprule[1.2pt]
			& \multicolumn{ 2}{c|}{Dehazing} & & \multicolumn{ 2}{c}{Deraining} \\
			Method &      SSIM &       PSNR & Method  &     SSIM &       PSNR				\\
			
			\hline
			\specialrule{0em}{1pt}{1pt}
			
			Baseline & 0.787 & 23.42 & Baseline   & 0.947 & 29.02 \\
			MSBDN    & 0.807 & 24.54 & Quan \etal & 0.945 & 28.23 \\
			Ours     & 0.806 & 24.93 & Ours       & 0.984 & 34.70 \\
			\bottomrule[1.2pt]
		\end{tabular}
	\end{center}
\end{table*}

\rev{To further show the generalization ability of our method, we have conducted additional experiments on dehazing and deraining. Visual results have been shown in Fig.~\ref{fig:dehaze_derain}. Numerical results are reported in Table~\ref{tab:app_LLV}. Here we train our model on standard RESIDE dataset~\cite{li2018benchmarking} for dehazing and PBRR dataset~\cite{Liu2019PBRR,Hao2019ICCVW} for deraining, respectively. We generate pseudo semantic labels via DETR~\cite{Semantic:carion2020end}. Here we conduct the recent dehazing method (MSBDN~\cite{dong2020multi}) and deraining approach (Quan \etal~\cite{quan2019deep}) as reference. Numerical results in Table~\ref{tab:app_LLV} demonstrate the effectiveness and generalization ability of our proposed framework.}

\end{document}